\documentclass{article}
\usepackage{xcolor}
\usepackage[most]{tcolorbox}
\usepackage[framemethod=tikz]{mdframed}
\usepackage{xparse}
\usepackage{tikz}

\newtcbox{\bracketeq}{
  nobeforeafter, enhanced, colback=gray!10, frame empty, math upper,
  boxsep=2pt, left=8pt, right=8pt, top=4pt, bottom=4pt, arc=4pt,
  tcbox raise base, 
  overlay={
    \def\thk{1.0pt}   
    \def\stub{4pt}    
    \draw[line width=\thk] (frame.north west) -- (frame.south west);
    \draw[line width=\thk] (frame.north east) -- (frame.south east);
    \draw[line width=\thk] (frame.north west) -- ([xshift=\stub]frame.north west);
    \draw[line width=\thk] (frame.south west) -- ([xshift=\stub]frame.south west);
    \draw[line width=\thk] (frame.north east) -- ([xshift=-\stub]frame.north east);
    \draw[line width=\thk] (frame.south east) -- ([xshift=-\stub]frame.south east);
  }
}

\newtcbox{\boxeq}{%
  enhanced,
  on line,
  nobeforeafter,
  math upper,          
  frame empty,
  colback=gray!10,
  boxsep=2pt, left=4pt, right=4pt, top=4pt, bottom=4pt,
  arc=3pt,
  tcbox raise base,
  overlay={
    \def\r{3pt}         
    \def\thk{1.0pt}     
    \def\stub{4pt}      
    \def\wipe{1pt}      

    \draw[line width=\thk, rounded corners=\r]
      (frame.north west) rectangle (frame.south east);

    \fill[white]
      ([xshift=\stub,yshift=\wipe]frame.north west)
      rectangle
      ([xshift=-\stub,yshift=-\wipe]frame.north east);
    \fill[white]
      ([xshift=\stub,yshift=\wipe]frame.south west)
      rectangle
      ([xshift=-\stub,yshift=-\wipe]frame.south east);
  }
}

\usepackage{PRIMEarxiv}

\usepackage[utf8]{inputenc} 
\usepackage[T1]{fontenc}    
\usepackage{hyperref}       
\usepackage{url}            
\usepackage{booktabs}       
\usepackage{amsfonts}       
\usepackage{nicefrac}       
\usepackage{microtype}      
\usepackage{lipsum}
\usepackage{fancyhdr}       
\usepackage{graphicx}       
\graphicspath{{media/}}     

\usepackage{xcolor}         
\definecolor{sidgray}{gray}{0.9}

\usepackage{natbib}
\usepackage{amsmath, amssymb, amsfonts, bm}
\usepackage{geometry}
\usepackage{wrapfig}
\usepackage{algorithm}
\usepackage{algorithmic}
\usepackage{booktabs}
\usepackage{multirow}
\usepackage{graphicx}
\usepackage{colortbl}
\usepackage{enumitem}
\usepackage{subcaption}
\usepackage{makecell}
  
\title{Simulation-Informed Diffusion for Decentralized Multi-robot Motion Planning
}

\author{
  Jinhao Liang \\
  University of Virginia \\
  \texttt{jliang@email.virginia.edu} \\
  \And
  Sven Koenig \\
  University of California, Irvine \\
  \texttt{sven.koenig@uci.edu} \\
  \And
  Ferdinando Fioretto \\
  University of Virginia \\
  \texttt{fioretto@virginia.edu} \\
}

\begin{document}
\maketitle

\begin{abstract}
Decentralized multi-robot motion planning requires each robot to generate collision-free trajectories from local observations, without global sensing or reliable communication. However, most existing planners, whether classical or learning-based, generate trajectories from a static snapshot of the local observation, which limits their ability to anticipate the future behavior of neighboring robots. This limitation is critical as the number of robots increases and the environment becomes more cluttered. Motivated by world models that use future state estimation to improve planning decisions, this paper introduces \emph{Simulation-Informed Diffusion} (SID), a decentralized framework built on constraint-aware diffusion models (CADM) that simulate neighboring robots' future trajectories to guide each robot's own planning. SID first uses CADM to simulate the future trajectories of neighboring robots from their currently observed states, and then uses the same CADM to plan each robot's own trajectory under safety constraints informed by these simulations. Crucially, the accurate simulation of neighbors enables a minimal communication scheme that triggers coordination only when necessary in highly congested scenarios. Experiments across diverse environments show that SID consistently outperforms baseline methods in terms of planning effectiveness and constraint satisfaction, and scales to scenarios with 108 robots and 160 obstacles.
\end{abstract}

\keywords{Diffusion models \and Decentralized multi-robot motion planning}

\section{Introduction}
Multi-Robot Motion Planning (MRMP) seeks collision-free trajectories for multiple robots operating in a shared environment, and underpins applications ranging from automated warehouses to drone fleets and autonomous driving systems~\citep{antonyshyn2023multiple}. Many MRMP methods assume a centralized planner with full access to all robot states~\citep{liang2025discrete,shaoul2024multi}. However, this assumption is often violated in practical systems, where communication may be limited or unreliable, and each robot may operate under partial observability of the environment~\citep {9384257}.
The key challenge in such settings is to generate collision-free trajectories from local observations with minimal communication.

A common way to address this decentralized planning problem is to compute feasible actions or trajectories based on the current local observation. In particular, \emph{reactive methods}, such as ORCA~\citep{van2011reciprocal}, select an immediate feasible action by reasoning over the current positions and velocities of nearby robots and obstacles. These methods are efficient and decentralized, but they are inherently myopic: a feasible action computed from a single snapshot ignores future interactions and may fail in dense environments. To mitigate this issue, \emph{learning-based methods} have been proposed to learn mappings from local observations to actions or future states~\citep{8661608,10342305}. These methods can capture more complex relationships between observations and actions, extending the reactive paradigm but inherit its myopia, since plans still derive from the current observation alone. The importance of anticipating future states for planning is also supported by recent work on world models~\citep{cen2025worldvla,li2026causal,ye2026world}.
Finally, \textit{prediction-based planning} methods predict future trajectories of neighboring robots and use these as planning references~\citep{9360433,10172259}. These methods provide forward-looking signals, but the limited accuracy of the predictions may cause substantial mismatch with the trajectories neighbors actually execute. This mismatch ultimately undermines the reliability of these references for decentralized planning.

\begin{figure}[t]
    \centering
    \includegraphics[width=\linewidth]{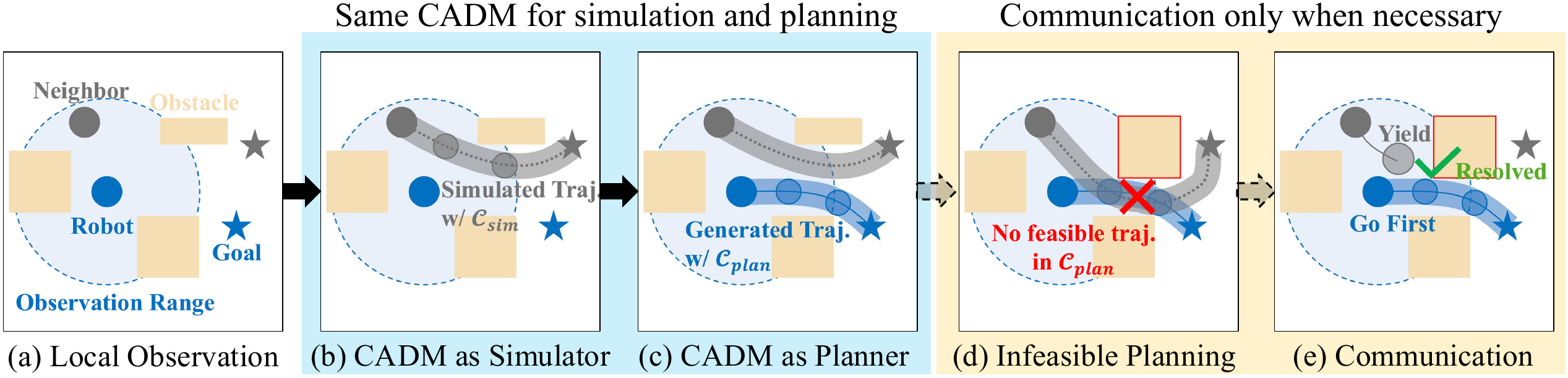}
    \caption{Overview of SID. From a local observation, the robot uses CADM first to simulate feasible future trajectories for neighboring robots and then to plan its own trajectory under safety constraints. If the local planner cannot find a feasible trajectory under the current simulated constraints, SID triggers communication among the affected robots to coordinate actions.
}
    \label{fig:sid_overview}
\end{figure}

To overcome these challenges, this paper introduces \textit{Simulation-Informed Diffusion (SID)}, a decentralized framework in which each robot uses a single \emph{constraint-aware diffusion model} (CADM) as both simulator and planner. The proposed constraint-aware diffusion model generates trajectories that satisfy collision-avoidance and kinematic constraints; it augments standard diffusion sampling with a proximal term and uses a Lagrangian dual method to solve the resulting constrained optimization problem.  
As a simulator, CADM generates accurate future trajectories for neighboring robots from their current observed states. As a planner, it combines these simulated trajectories with the current local observation to generate a collision-free trajectory by projecting diffusion samples onto the planning constraint set. Crucially, the accuracy of this constraint-aware diffusion-based simulator enables each robot to greatly minimize communication. The proposed communication scheme triggers messages only when local planning cannot resolve a conflict, eliminating unnecessary exchange in sparse conditions. 
In particular, SID develops a communication mechanism driven by simulation-informed conflict detection. Robots exchange messages only when local simulation and planning cannot resolve predicted conflicts, reducing communication while maintaining planning performance. In an extensive MRMP benchmark of increasing size and complexity, SID outperforms representative reactive, learning-based, and prediction-based planning methods (ORCA~\citep{van2011reciprocal}, IA-MPC~\citep{9360433}, and MIMIC-D~\citep{dong2025mimic}) in trajectory quality and constraint satisfaction, and in large-scale congested scenarios it remains effective using up to 108 robots and 160 obstacles. Figure~\ref{fig:sid_overview} summarizes the SID framework.

\section{Related Work}
\noindent \textbf{Multi-robot Motion Planning.}
Multi-robot motion planning has traditionally been formulated as a joint planning problem, where a centralized planner computes collision-free trajectories for all robots. Classical approaches typically instantiate this view through optimization-based formulations~\citep{6385823,marcucci2024shortest,jaitly2025milp}. These methods can reason about global feasibility and robot-robot interactions, but they require a central planner that collects all robot states, solves a coupled planning problem, and distributes commands. This creates a single point of failure~\citep{sheng2006distributed} and requires reliable communication, which is difficult to maintain in distributed environments~\citep{6942705}. These limitations motivated decentralized MRMP, where each robot plans from local observations and limited communication. In such a setting, \emph{reactive collision-avoidance methods}~\citep{van2011reciprocal,9384148,alonso2013optimal} removed the need for a central planner by allowing each robot to select locally safe actions from nearby robot states and obstacles; however, given their myopic nature, they often fail in dense environments where future interactions determine safety. \emph{Learning-based decentralized planners} extend this local decision-making paradigm by learning policies or planning rules from local observations~\citep{8661608,9366340,10342305}. These methods improve expressiveness over hand-designed reactive rules, but many remain tied to snapshot-based inputs~\citep{pmlr-v155-wang21d}. More recent methods predict future trajectories of neighboring robots and use them as planning references~\citep{9360433,10172259}.  This line of work introduces an important capability: local planners can reason about where nearby robots may be over the planning horizon, rather than only where they are now. However, standard neural predictors are not themselves constrained planners. As a result, they can produce trajectories that are dynamically infeasible, nonsmooth, or inconsistent with the trajectories that neighboring robots will actually execute. SID addresses this limitation by replacing unconstrained prediction with constraint-aware simulation, generating feasible future references that are aligned with the planner's trajectory distribution.

\noindent \textbf{Generative Models for Motion Planning.}
Generative models have recently provided a different route for motion planning: instead of searching for a single trajectory under a hand-designed cost, they learn distributions over feasible or high-quality trajectories. This is particularly attractive in continuous motion planning, where the solution space is multimodal and where a robot may need to choose among qualitatively different homotopy classes. Such approaches have been applied to single-robot motion planning~\citep{11097366, luo2024potential,ma2026drift,ma2026posterior} and centralized MRMP~\citep{luan2025projected, shaoul2024multi}, with projection and discrete guidance improving safety and scalability~\citep{liang2025simultaneous, liang2025discrete}. Recent work has also explored decentralized MRMP. 
However, they either condition each robot only on local observations~\citep{zhu2024madiff,dong2025mimic}, which limits their ability to anticipate future neighbor behavior, or require fixed-frequency communication among robots~\citep{lew2025aid}, which increases communication burden. SID closes this gap by using the same constraint-aware diffusion model as a local simulator and planner. Each robot simulates feasible future trajectories for its neighbors, conditions its own planning on these simulated futures, and communicates only when simulation-informed planning cannot resolve a predicted conflict.

\section{Preliminaries}
The paper focuses on decentralized multi-robot motion planning under partial observability and limited communication. We first introduce the problem formulation, then review generative diffusion models, which are the foundation of our proposed method.

\textbf{Decentralized multi-robot motion planning.}
Consider a set $\mathcal{A}=\{1,\dots,N\}$ of $N$ robotic agents operating within an environment $\mathcal{W}\!\subset\!\mathbb{R}^2$. The environment contains a set of static obstacles $\mathcal{O}$, defining the \emph{free space} as $\mathcal{F} := \mathcal{W}\!\setminus\! \mathcal{O}$.
Each robot $i\in\mathcal{A}$ starts from an initial state $\mathbf{s}_i$ and aims to reach a goal state $\mathbf{g}_i$. 
Unlike centralized MRMP, where a planner observes the full joint state of all robots, in the decentralized setting each robot plans from local information alone. Robot $i$ receives a local observation $\mathbf{o}_i$ encoding the portion of the environment and neighboring robots visible within its sensing range~\citep{9384257}. When communication is available, robot $i$ may also receive messages $\mathbf{c}_i = \{ \mathbf{m}_{j \to i} \}_{j \in \mathcal{N}_i}$ from neighboring robots in its communication neighborhood $\mathcal{N}_i$~\citep{10342305}.
The decentralized planner $\pi_\theta$ aims to map these local inputs to a collision-free trajectory of horizon $H$. This planned trajectory for robot $i$ is denoted as $\bm{\tau}_i \!=\! (\mathbf{p}_i^{0}, \mathbf{p}_i^{1}, \dots, \mathbf{p}_i^{H})$. 
The planner is defined as:
\begin{equation}
    \bm{\tau}_i = \pi_\theta(\mathbf{o}_i, \mathbf{c}_i, \mathbf{g}_i).
\end{equation}

Following a standard convention in MRMP, this work models each robot $i$ as a disk of radius $r_i$ and assumes a maximum speed $v_{\max}$~\citep{shaoul2024multi,liang2025simultaneous}. A set of trajectories $\{\bm{\tau}_i\}_{i\in\mathcal{A}}$ is feasible if it satisfies obstacle avoidance, inter-robot separation, and kinematic constraints at every planning step:
\begin{equation}
\label{eq:constraints}
    \begin{aligned}
    &\mathrm{dist}(\mathbf{p}_i^{\psi},\mathcal O)\ge r_i,
        \quad \forall i\in\mathcal A,\ \psi \in [H],\\
    &\|\mathbf{p}_i^{\psi} - \mathbf{p}_j^{\psi}\|_2 \ge r_i+r_j;
        \quad \forall i,j\in\mathcal A,\ i\neq j,\ \psi \in [H],\\
    &\|\mathbf{p}_i^{\psi+1}-\mathbf{p}_i^{\psi}\|_2 \le v_{\max},
        \quad \forall i\in\mathcal A,\ \psi \in [H{-}1],
    \end{aligned}
\end{equation}
where $\psi$ denotes the planning step. The first constraint requires each robot to remain in the obstacle-free region with clearance at least its radius. The second enforces pairwise separation between robots. The third imposes a maximum displacement between consecutive planning steps, which acts as a speed limit. SID uses these constraints both to guide trajectory generation and to determine when local simulation is insufficient and communication is required.

\textbf{Generative diffusion models for trajectory planning.}
Generative diffusion models \cite{ma2026learning, ho2020denoising} produce high-fidelity complex data by operating in two phases: a \emph{forward step}, that gradually introduces noise to clean data samples, followed by a \emph{reverse denoising step}, in which a deep neural network is trained to iteratively remove noise. This forward process defines a Markov chain $\{\bm{x}_t\}_{t=0}^T$, with initial sample $\bm{x}_0 \!\sim\! p(\bm{x}_0)$ and each transition \( q(\bm{x}_t | \bm{x}_{t-1}) \) realized by adding Gaussian noise according to a variance schedule $\alpha_t$. In trajectory planning, the clean sample is a trajectory over a finite horizon, e.g., $\bm{x}_0 \!\equiv\! \bm{\tau}_i \!=\! (\mathbf{p}_i^0,\mathbf{p}_i^1,\ldots,\mathbf{p}_i^H)$, or a vectorized representation of it. 
In \emph{score-based diffusion} \cite{song2019generative, song2020score}, the reverse process relies on a learned score function \(\bm{s}_{\theta}(\bm{x}_t, t) \!=\! \nabla_{\bm{x}_t}\! \log p_t(\bm{x}_t)\) which points toward regions of higher probability under the noisy distribution.
This score function is trained to minimize
\begin{equation}
\label{eq:score}
    \displaystyle 
    \min_{\theta} \mathbb{E}_{\substack{t \sim [1,T], p(\bm{x}_0),\\ 
    q(\bm{x}_t|\bm{x}_0)}}
    (1 -{\alpha}_t)
    \left[ \left\| \bm{s}_{\theta}(\bm{x}_t, t) - \nabla_{\bm{x}_t} \log q(\bm{x}_t|\bm{x}_0) \right\|^2 \right].
\end{equation}
where $\alpha_t$ is a variance schedule that controls the noise level at each step.  Once trained, the model generates a trajectory by initializing $\bm{x}_T\sim\mathcal{N}(\mathbf{0},\mathbf{I})$ and iteratively denoising it. 
This is also known as the \emph{sampling} phase, which commonly follows the Stochastic Gradient Langevin Dynamics update rule~\citep{song2019generative,ma2026transition}:
\begin{equation}
    \bm{x}_t \!=\! \bm{x}_{t+1} + \frac{\epsilon}{2} \bm{s}_\theta(\bm{x}_{t+1}, t\!+\!1) \!+\! \sqrt{\epsilon} \bm{z}, \label{eq:sgld}
\end{equation}
where $\epsilon$ denotes the step size and $\bm{z}$ is Gaussian noise.

\section{Simulation-Informed Diffusion for Decentralized MRMP}
This section presents Simulation-Informed Diffusion (SID), a diffusion-based framework for decentralized MRMP. We first describe the Constraint-aware Diffusion Model (CADM), which generates feasible trajectories under safety constraints. We then use CADM as a local simulator to predict feasible future trajectories for neighboring robots. Finally, we present a simulation-guided diffusion planner that conditions each robot's trajectory generation on these simulated futures, together with a minimal communication mechanism that activates only when local planning becomes infeasible.

\textbf{Constraint-aware Diffusion Models.}
Diffusion models can capture complex trajectory distributions, but they may still produce infeasible trajectories, limiting their use in robotic systems~\citep{11097366,liang2025simultaneous,liang2026improved}. 
To enforce collision-avoidance and kinematic constraints during sampling, we build on projected diffusion methods for motion planning~\citep{Fioretto:NeurIPS24,liang2025simultaneous} and define the constrained target density at diffusion step $t$ as $\tilde{p}_t(\bm{x}_t) \propto p_t(\bm{x}_t)\mathbb{I}(\bm{x}_t \in \mathcal{C})$, where $\mathcal{C}$ denotes the feasible set induced by the planning constraints. Unlike standard diffusion sampling, which seeks only high-probability samples under $p_t$, this formulation seeks high-probability trajectories while requiring every diffusion iterate to lie in $\mathcal{C}$. 
Operationally, CADM modifies the SGLD update in Eq.~\ref{eq:sgld} to include a projection step:
\begin{equation}
    \bm{x}_t \!=\! \mathcal{P}_{\mathcal{C}}\!\left(\bm{x}_{t+1} + \frac{\epsilon}{2} \bm{s}_\theta(\bm{x}_{t+1}, t\!+\!1) \!+\! \sqrt{\epsilon} \bm{z}\right), \label{eq:projected_sgld}
\end{equation}
where $\mathcal{P}_{\mathcal{C}}(\bm{x}_t) = \arg\min_{\bm{y}\in \mathcal{C}} \|\bm{y} - \bm{x}_t\|_2^2$ is the Euclidean projection operator onto the constraint set $\mathcal{C}$. This projection enforces feasibility at every sampling step and guides the diffusion model toward constraint-satisfying trajectories.

Given the non-convexity of $\mathcal{C}$, this projection is non-trivial. However, the structure of the constraints allows for efficient optimization. The obstacle avoidance and inter-robot separation constraints can be expressed as differentiable functions of the trajectory, while the kinematic constraints are simple norm bounds on consecutive trajectory points. We write $\mathcal{C} = \{\bm{y} \mid \mathbf{h}(\bm{y}) \le \mathbf{0}\}$, where $\mathbf{h}(\bm{y})$ stacks the collision-avoidance and kinematic inequality constraints, and solve the resulting projection problem using the augmented Lagrangian:
\begin{equation}
    \mathcal{L}_{\rho}(\bm{y}, \boldsymbol{\nu}) = \|\bm{y} - \bm{x}_t\|_2^2 + \boldsymbol{\nu}^\top \mathbf{h}(\bm{y}) + \frac{\rho}{2} \|\max(\mathbf{0}, \mathbf{h}(\bm{y}))\|_2^2,
\end{equation}
where $\boldsymbol{\nu} \ge \mathbf{0}$ are the Lagrange multipliers and $\rho > 0$ is the penalty parameter. The projection sub-problem is solved by leveraging the classical primal-dual optimization method: we alternate gradient descent on the primal variable $\bm{y}$ and dual ascent on $\boldsymbol{\nu}$, with update $\boldsymbol{\nu} \leftarrow \max(\mathbf{0}, \boldsymbol{\nu} + \rho \, \mathbf{h}(\bm{y}))$~\citep{neal2011distributed,boyd2004convex}. After each round, $\rho$ is increased. Projection stops when the maximum constraint violation falls below $10^{-5}$ or after 100 primal--dual iterations. If projection fails at an intermediate denoising step, we retain the minimum-violation iterate and continue sampling. If the final trajectory remains infeasible, we identify the bottleneck set and trigger communication-based replanning.

\textbf{Neighbor Trajectory Simulation.}
\label{sec:neighbor_simulation}
Even if trajectory feasibility can be enforced by the CADM sampler, in decentralized MRMP, the inter-robot constraints in Eq.~\ref{eq:constraints} depend on the future positions of neighboring robots over the planning horizon. If these future positions are unavailable, the planning robot can only constrain its trajectory against the currently observed neighbor states, which reduces CADM to a reactive collision-avoidance planner. SID avoids this limitation by using CADM not only to plan the agent's trajectory, but also to simulate feasible future trajectories for observed neighbors. These simulated trajectories define the time-indexed references against which each robot enforces inter-robot safety. Formally, robot $i$ applies the simulator $\pi_\theta$ independently to each observed neighbor:
\begin{equation}
    \hat{\bm{\tau}}_j = \pi_\theta(\mathbf{o}_i, \mathbf{c}_i, \mathbf{g}_j), \quad \forall j \in \mathcal{N}_i,
\end{equation}
where $\hat{\bm{\tau}}_j = (\hat{\mathbf{p}}_j^{0}, \hat{\mathbf{p}}_j^{1}, \dots, \hat{\mathbf{p}}_j^{H})$ denotes the simulated trajectory of neighbor $j$ over the planning horizon $H$. We assume neighboring goals $\mathbf{g}_j$ are provided at initialization. Thus, the neighboring goal $\mathbf{g}_j$ is known and does not require repeated online communication. The key role of this step is not merely to predict where neighbor $j$ may move, but to produce a realistic trajectory reference generated by the same constraint-aware diffusion mechanism used for planning, as shown in Figure~\ref{fig:sid_overview}(b).

To serve as a reliable planning reference, $\hat{\bm{\tau}}_j$ must satisfy obstacle avoidance, kinematic limits, and inter-robot separation. When simulating neighbor $j$, robot $i$ treats the stored planned trajectories of the other observed neighbors from the previous planning round as fixed references; denote these trajectories by $\bm{\tau}_l = (\mathbf{p}_l^0,\ldots,\mathbf{p}_l^H)$. The resulting simulation constraint set is:
\begin{equation}
    \mathcal{C}_{\mathrm{sim}}(j)
    =
    \left\{
    \hat{\bm{\tau}}_j
    \,\middle|\,
    \begin{aligned}
    &\mathrm{dist}(\hat{\mathbf{p}}_j^{h},\mathcal{O}) \ge r_j,
    && h\in[H], \\
    &\|\hat{\mathbf{p}}_j^{h+1}-\hat{\mathbf{p}}_j^{h}\|_2 \le v_{\max},
    && h\in[H{-}1], \\
    &\|\hat{\mathbf{p}}_j^{h}-\mathbf{p}_l^{h}\|_2 \ge r_j+r_l,
    && \forall l \in \mathcal{N}_i\setminus\{j\},\ h\in[H]
    \end{aligned}
    \right\}.
    \label{eq:simulation_constraint_set}
\end{equation}

\textbf{Simulation-guided Diffusion-based Decentralized Planning.}
Given the simulated neighboring trajectories $\{\hat{\bm{\tau}}_j\}_{j \in \mathcal{N}_i}$, robot $i$ uses them as dynamic references for planning its trajectory, which is illustrated in Figure~\ref{fig:sid_overview}(c).  We adopt a receding-horizon control (RHC) scheme for decentralized planning~\cite{zhu2024madiff,dong2025mimic}. At planning time $\psi$, robot $i$ acquires a local observation $\mathbf{o}_i^{\psi}$, simulates the future trajectories of observed neighbors, and then generates a trajectory $\bm{\tau}_i = (\mathbf{p}_i^{0}, \mathbf{p}_i^{1}, \dots, \mathbf{p}_i^{H})$. Although the planner optimizes over all $H$ future steps, it executes only the first $k$, where $1 \le k < H$. After execution, robot $i$ receives a new local observation $\mathbf{o}_i^{\psi+k}$, re-simulates the neighboring trajectories $\{\hat{\bm{\tau}}_j^{\psi+k}\}_{j \in \mathcal{N}_i^{\psi+k}}$, and replans over the next horizon. This observation-simulation-planning-execution loop lets the local plan adapt as interactions evolve.

To guarantee safety in each planning round, the planned trajectory for robot $i$ must satisfy environmental obstacle avoidance, kinematic limits, and inter-robot safety with respect to the simulated neighboring trajectories. The planning constraint set can be written as:
\begin{equation}
    \mathcal{C}_{\text{plan}}(i)
    =
    \left\{
    \bm{\tau}_i
    \,\middle|\,
    \begin{aligned}
    &\mathrm{dist}(\mathbf{p}_i^{h},\mathcal{O}) \ge r_i,
    \quad h\in[H], \\
    &\|\mathbf{p}_i^{h+1}-\mathbf{p}_i^{h}\|_2 \le v_{\max},
    \quad h\in[H{-}1], \\
    &\|\mathbf{p}_i^{h}-\hat{\mathbf{p}}_j^{h}\|_2 \ge r_i+r_j,\
    \forall j \in \mathcal{N}_i,\ h\in[H]
    \end{aligned}
    \right\}.
\end{equation}
\begin{wrapfigure}[16]{r}{0.58\columnwidth} 
\vspace{-15pt}
    \begin{minipage}{\linewidth} 
        \begin{algorithm}[H]
            \caption{Simulation-guided Decentralized Planning}
            \label{alg:Simulation-guided-Decentralized-Planning}
            \begin{algorithmic}[1]
            \STATE \textbf{Input:} Robot $i$, goal $\mathbf{g}_i$, CADM $\pi_\theta$, RHC window $k$
            \STATE Initialize planning time $\psi \leftarrow 0$
            \WHILE{robot $i$ not at $\mathbf{g}_i$}
                \STATE Acquire $\mathbf{o}_i$ and observed neighbors $\mathcal{N}_i$
                \FOR{each neighbor $j \in \mathcal{N}_i$}
                    \STATE Simulate $\hat{\bm{\tau}}_j$ with $\pi_\theta$ under $\mathcal{C}_{\text{sim}}(j)$
                \ENDFOR
                \STATE Define ${\mathcal{C}}_{\text{plan}}(i)$ and $\widetilde{\mathcal{C}}_{\text{plan}}(i)$ using $\{\hat{\bm{\tau}}_j\}_{j \in \mathcal{N}_i}$
                \STATE Define $\mathcal{C}_{\text{hold}}(i)$ using space-time A$^\ast$ on $\widetilde{\mathcal{C}}_{\text{plan}}(i)$
                \STATE Generate $\bm{\tau}_i$ with $\pi_\theta$ under $\mathcal{C}_{\text{plan}}(i)\cap \mathcal{C}_{\text{hold}}(i)$
                \STATE Execute the first $k$ time steps of $\bm{\tau}_i$
                \STATE Advance planning time $\psi \leftarrow \psi + k$
            \ENDWHILE
            \end{algorithmic}
            \end{algorithm}
    \end{minipage}
\end{wrapfigure}
The set $\mathcal{C}_{\text{plan}}(i)$ enforces feasibility with respect to environmental obstacles and simulated neighboring trajectories, but it does not determine when robot $i$ should reach its goal. In diffusion-based planning, the travel time is typically predefined because the planner samples trajectories over a fixed horizon~\citep{shaoul2024multi,liang2025simultaneous}. However, a fixed arrival schedule is restrictive in decentralized MRMP: robots in sparse regions should reach their goals sooner, whereas robots in congested regions may need more time to avoid neighbors safely.

To address this issue, we estimate an adaptive arrival time for each robot at each RHC round and impose it as a goal-holding constraint, allowing the planner to reduce travel time while preserving safe goal reaching. Specifically, at each planning round, robot $i$ runs a space-time A$^\ast$ search on a discretized version of the local planning problem under $\mathcal{C}_{\text{plan}}(i)$~\citep{silver2005cooperative}. Let $\widetilde{\mathcal{C}}_{\text{plan}}(i)$ denote the resulting discretized feasible set. The search returns the earliest feasible arrival step
$
a_i = \min \{ h \in \{0,\dots,H\} \mid \tilde{\mathbf{p}}_i^{h}=\mathbf{g}_i \}.
$
Because space-time A$^\ast$ operates on a discretized feasible set, $a_i$ may be later than the shortest feasible arrival time in the continuous trajectory space used by SID. We therefore treat $a_i$ as a conservative arrival step. Given $a_i$, we define goal-holding constraints
\begin{equation}
    \mathcal{C}_{\text{hold}}(i)
    =
    \left\{
    \bm{\tau}_i
    \,\middle|\,
    \mathbf{p}_i^{h}=\mathbf{g}_i,\quad
    h=a_i,\dots,H
    \right\}.
\end{equation}
Overall, robot $i$ plans under the combined feasible set $\mathcal{C}_{\text{plan}}(i)\cap \mathcal{C}_{\text{hold}}(i)$. This combined constraint keeps the trajectory safe with respect to obstacles and simulated neighbors while ensuring that it reaches the goal by the conservative target step and remains there afterward.
The complete procedure is summarized in Algorithm~\ref{alg:Simulation-guided-Decentralized-Planning}.

\textbf{Communication Mechanism Design.} 
Simulation-guided planning resolves most local interactions by using 
\begin{wrapfigure}[13]{r}{0.53\columnwidth}
    \vspace{-8pt}
    \centering
    \includegraphics[width=\linewidth]{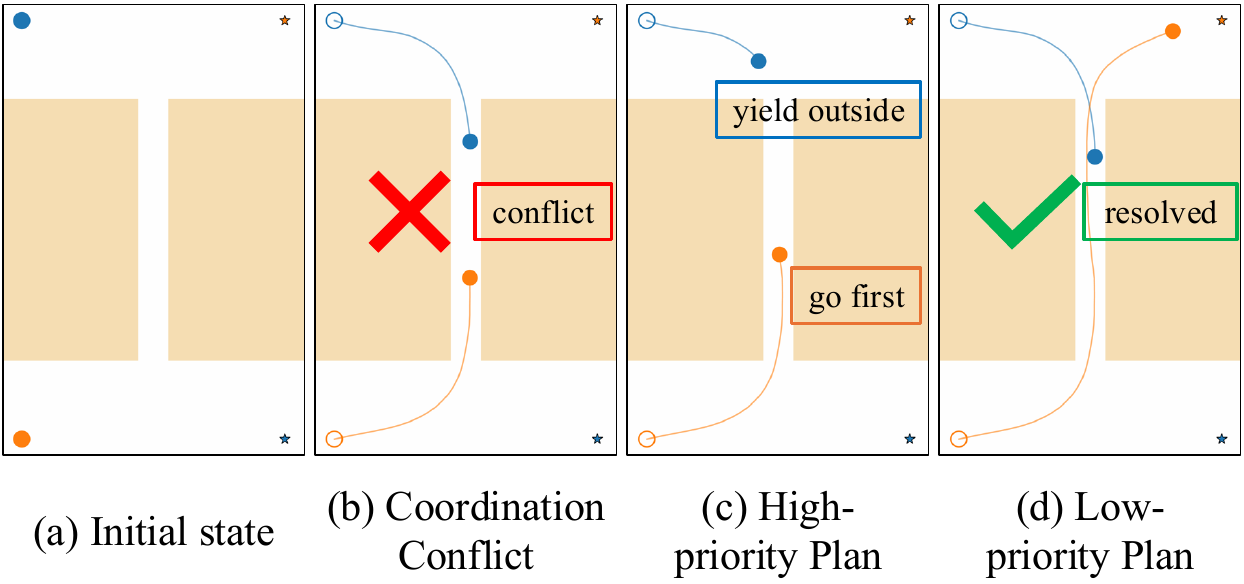}
    \caption{Example of a narrow corridor bottleneck.}
    \label{fig:communication}
\end{wrapfigure}
simulated future trajectories as references for local trajectory generation.
However, some crowded interactions require explicit coordination rather than simulation alone, as shown in Figure~\ref{fig:sid_overview}(d)--(e). Consider a case in which several robots may try to enter a narrow passage that admits only one robot at a time. Local observations and simulated trajectories alone cannot determine a passing order, as illustrated in Figure~\ref{fig:communication}, panels (a) and (b). In such cases, robots need communication to share planned trajectories and break the symmetry of the interaction.
To address this issue while minimizing communication, we add a minimal communication mechanism on top of the simulation-guided planner. 
Accurate simulation and simulation-guided planning give SID a concrete trigger for intervention: if Algorithm~\ref{alg:Simulation-guided-Decentralized-Planning} cannot find a feasible trajectory under $\mathcal{C}_{\text{plan}}(i)$, then robot $i$ has reached a coordination bottleneck that local simulation alone cannot resolve.

\begin{algorithm}[t]
\caption{Minimal Communication Mechanism}
\label{alg:comm-planning}
\begin{algorithmic}[1]
\STATE \textbf{Input:} Robot $i$, planning constraint $\mathcal{C}_{\text{plan}}(i)$, neighbors $\mathcal{N}_i$
\STATE Identify bottleneck set $\mathcal{B}_i \subseteq \mathcal{N}_i \cup \{i\}$
\STATE Compute $d_j = \|\mathbf{p}_j - \mathbf{g}_j\|_2$ for all $j \in \mathcal{B}_i$
\STATE Rank robots in $\mathcal{B}_i$ by ascending $d_j$
\STATE Remove inter-robot constraints involving $\mathcal{B}_i \setminus \{i\}$ from $\mathcal{C}_{\text{plan}}(i)$
\IF{$i$ is not first in this order}
    \STATE Receive one trajectory $\bm{\tau}_j$ from each earlier robot $j$
    \STATE Add $\bm{\tau}_j$ to $\mathcal{C}_{\text{plan}}(i)$ for each earlier robot $j$
\ENDIF
\STATE Generate $\bm{\tau}_i$ under the updated $\mathcal{C}_{\text{plan}}(i)$
\STATE Broadcast $\bm{\tau}_i$ once to later robots in $\mathcal{B}_i$
\STATE \textbf{return} $\bm{\tau}_i$
\end{algorithmic}
\end{algorithm}

The mechanism therefore remains inactive during routine planning and activates only when local planning becomes infeasible. When this trigger fires, robot $i$ first identifies the robots involved in the unresolved local interaction. We denote this bottleneck set by $\mathcal{B}_i$. Specifically, the set $\mathcal{B}_i$ contains robot $i$ and observed neighbors whose inter-robot collision-avoidance residual exceeds $10^{-5}$ at any step of the planning horizon. SID then assigns priority within $\mathcal{B}_i$ using each robot's remaining distance to goal, $d_j = \|\mathbf{p}_j - \mathbf{g}_j\|_2$ for $j \in \mathcal{B}_i$.
Robots with smaller $d_j$ receive higher priority, which yields a deterministic passing order, as illustrated in Figure~\ref{fig:communication}, panels (c) and (d).
The fallback planner replans robots in $\mathcal{B}_i$ according to this priority order, as shown in Algorithm~\ref{alg:comm-planning}. Before replanning, each robot removes from its planning constraint set the inter-robot constraints associated with robots in $\mathcal{B}_i \setminus \{i\}$. The first robot in the order therefore plans without constraints from later robots and broadcasts its trajectory once. Each later robot waits for the planned trajectories of all earlier robots. For each earlier robot $j$, the current robot adds the communicated trajectory $\bm{\tau}_j$ to its planning constraint set and replans under the updated constraints. After replanning, it broadcasts its own trajectory to later robots in $\mathcal{B}_i$. Thus, communication is restricted to the bottleneck set, and each robot transmits at most one trajectory during a fallback episode. The communicated trajectories convert an ambiguous simultaneous interaction into an ordered planning problem, resolving the symmetry that caused planning infeasibility.

\section{Experiments}
\begin{figure}[t]
    \centering
    \begin{subfigure}[t]{0.2\linewidth}
        \centering
        \includegraphics[width=\linewidth]{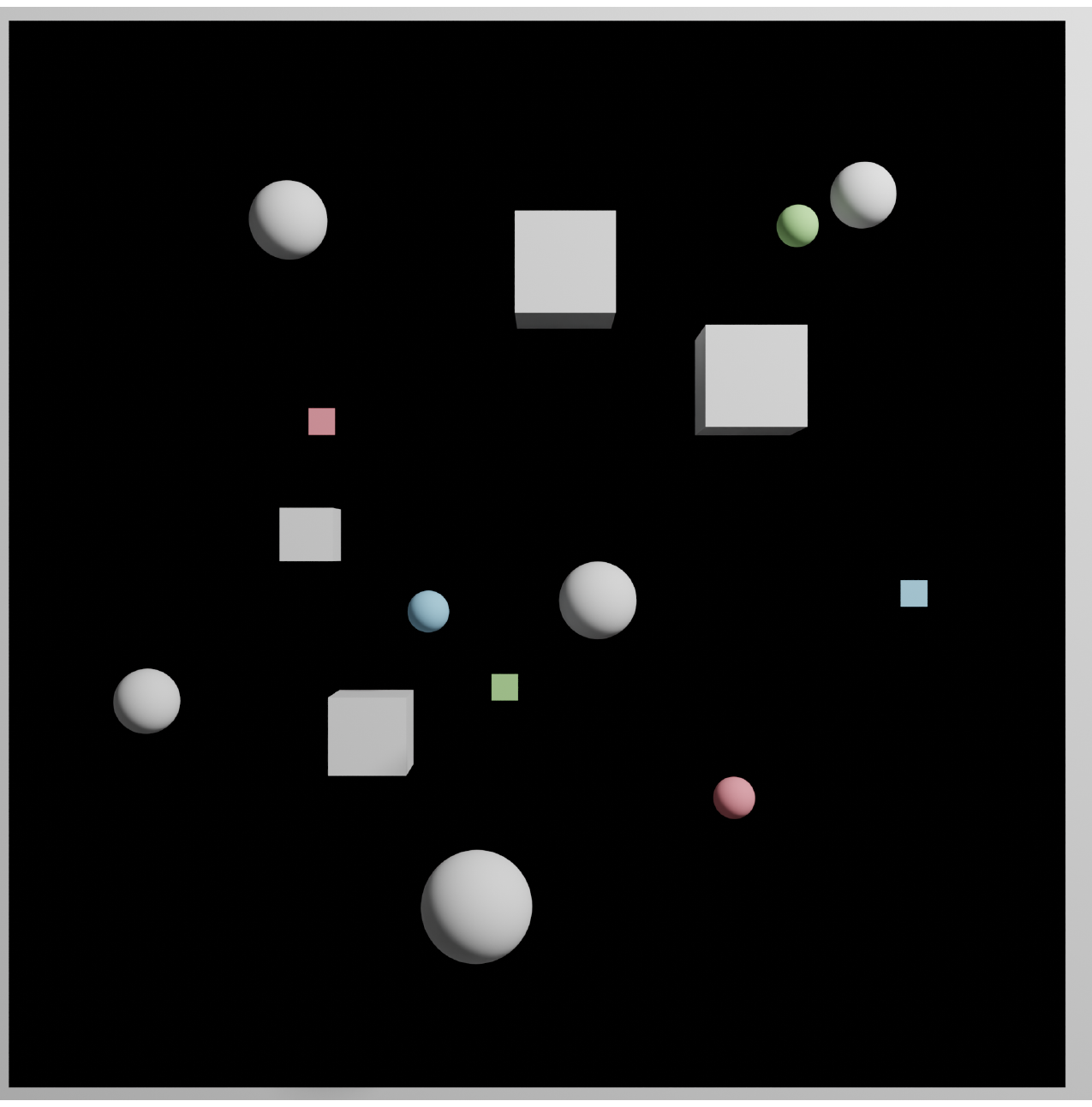}
        \caption{Basic.}
    \end{subfigure}
    \hfill
    \begin{subfigure}[t]{0.2\linewidth}
        \centering
        \includegraphics[width=\linewidth]{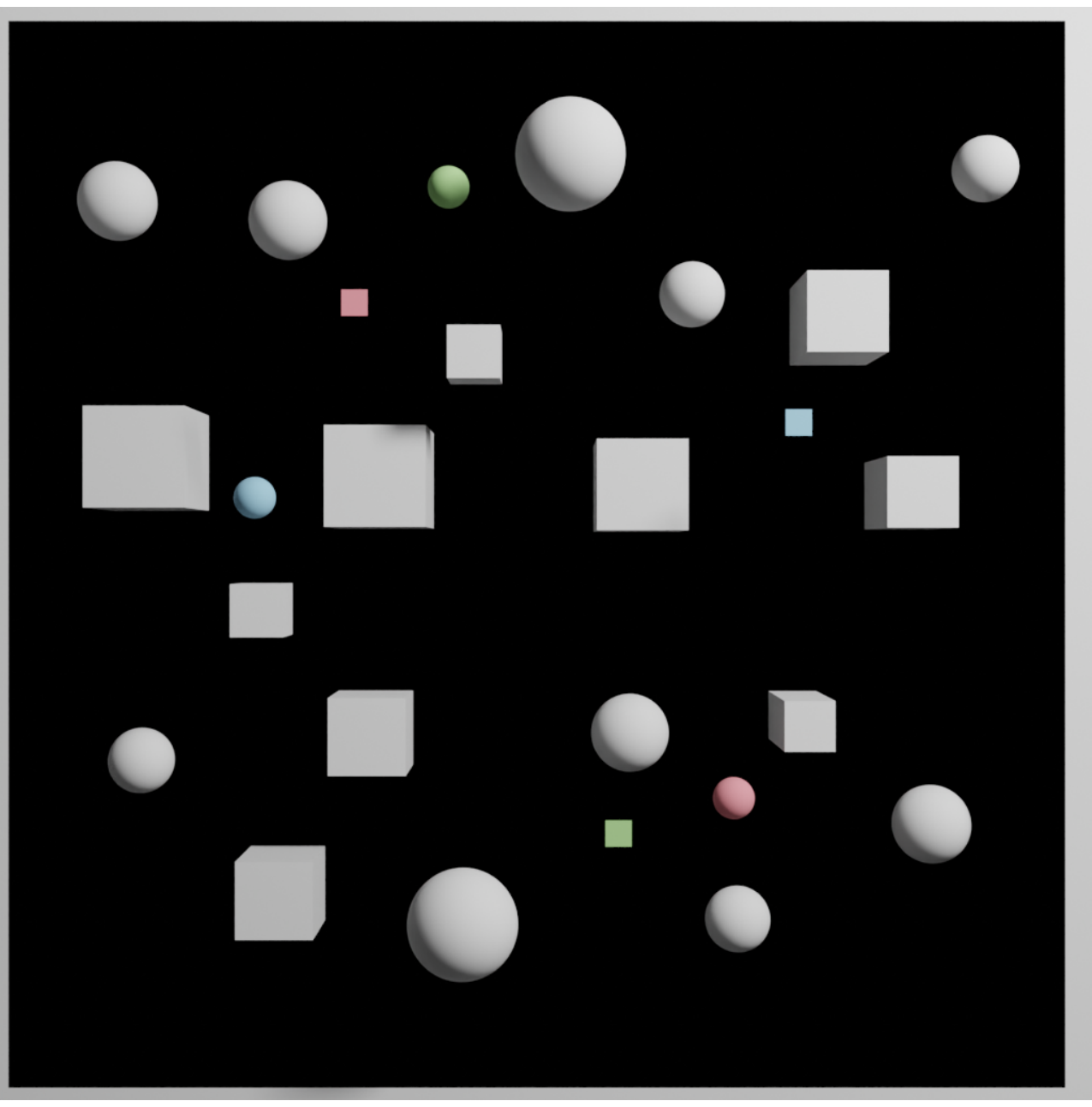}
        \caption{Dense.}
    \end{subfigure}
    \hfill
    \begin{subfigure}[t]{0.2\linewidth}
        \centering
        \includegraphics[width=\linewidth]{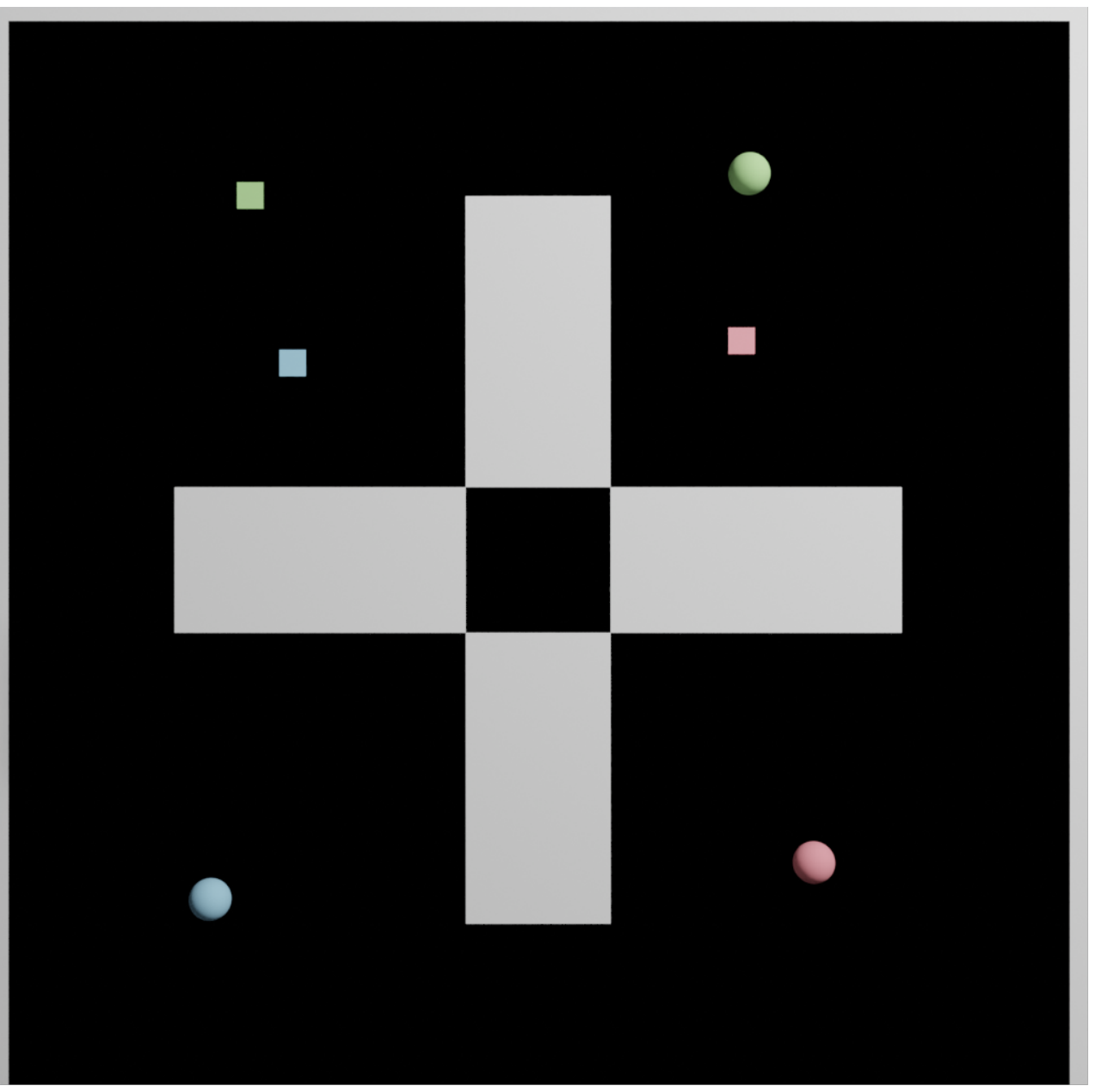}
        \caption{Room.}
    \end{subfigure}
    \hfill
    \begin{subfigure}[t]{0.2\linewidth}
        \centering
        \includegraphics[width=\linewidth]{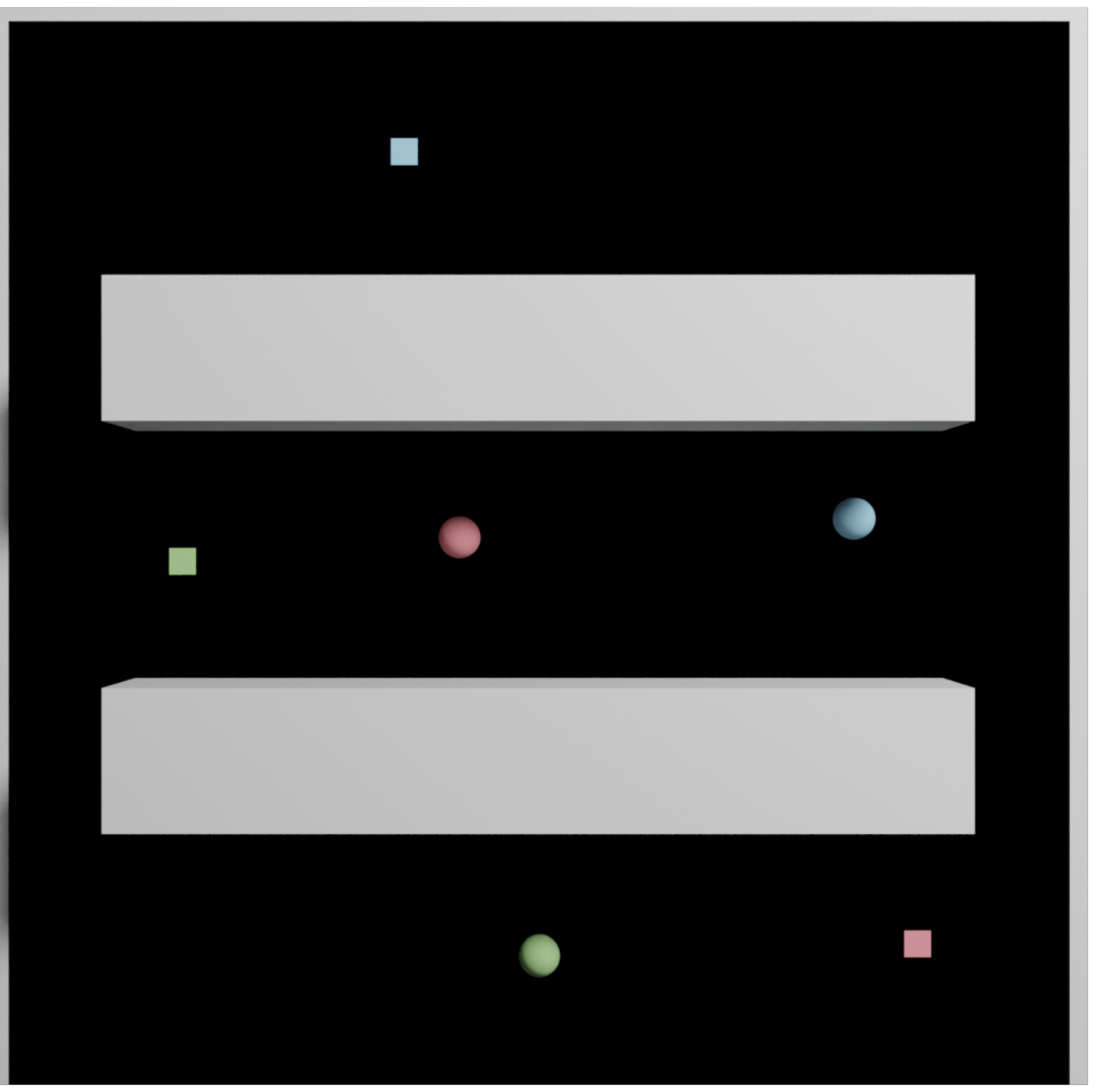}
        \caption{Shelf.}
    \end{subfigure}
    \caption{Examples of benchmark maps used for MRMP experiments.}
    \label{fig:benchmark_maps}
\end{figure}

\textbf{Settings.}
This paper evaluates decentralized MRMP algorithms in two complementary settings. \textbf{(1)} The first setting tests \textbf{Feasibility, Scalability, and Quality} on a modified MRMP benchmark adapted from~\cite{liang2025simultaneous}. Each map has spatial extent $2\times2$. The benchmark includes four representative map types: random maps with varying obstacle densities (basic and dense), and structured maps that mimic real-world environments (room and shelf layouts), as shown in Figure~\ref{fig:benchmark_maps}.
Colorful spheres and plates denote the starts and goals of robots, and white objects indicate obstacles. We evaluate each map type with 3, 6, and 9 robots across 25 different configurations. To test scalability beyond the original benchmark, we further extend the basic and dense maps to larger teams with 12, 15, and 18 robots.
The local observation range is measured as a fraction of the map size and is set to $0.5$ by default. The receding-horizon controller executes $k=8$ steps at each planning round, and the diffusion model outputs trajectories with a maximum length of 64 steps. We further analyze sensitivity to the observation range and execution window. \textbf{(2)} The second setting evaluates SID on \textbf{challenging scenarios}. These scenarios test scalability under severe congestion, including large-scale instances with 108 robots and 160 obstacles.

\textbf{Evaluation Metrics.}
We evaluate each algorithm using \underline{Success rate}, \underline{Communication Frequency}, \underline{Smoothness}, and \underline{Travel time}. A test case is successful only if every robot reaches its goal and no collision occurs at any time. \emph{Success rate} is the percentage of successful test cases. \emph{Communication Frequency} is the total number of communication events in an instance divided by the number of robots, averaged over test instances. \emph{Smoothness} is computed from the squared discrete acceleration along each trajectory and reported as the mean per-agent accumulated value. \emph{Travel time} is the mean arrival time across all robots in a successful instance. \emph{Smoothness} and \emph{Travel time} are computed only on successful test cases.

\textbf{Competing Methods.}
We compare SID with three competitive decentralized baselines that cover the main families of multi-robot coordination methods:
\begin{enumerate}[leftmargin=*, parsep=0pt, itemsep=0pt, topsep=0pt]
\item \textbf{Optimal Reciprocal Collision Avoidance (ORCA)} is a classical decentralized reactive planner that selects collision-avoiding velocities under reciprocal responsibility assumptions~\citep{van2011reciprocal}.
\item \textbf{Interaction-Aware Model Predictive Control (IA-MPC)} combines learned interaction-aware trajectory prediction with model predictive control, providing a predictive optimization baseline for dynamic multi-robot interactions~\citep{9360433}.
\item \textbf{Multi-agent Coordination with Decentralized Diffusion Policies (MIMIC-D)} learns decentralized diffusion policies for multi-agent coordination through multimodal imitation, providing a state-of-the-art generative planning baseline~\citep{dong2025mimic}.
\end{enumerate}
All methods use the same observation range and execution window during evaluation. All neural-network-based methods are trained on the same dataset. The details of the implementation, additional results, and visualizations for challenging scenarios are provided in Appendices~\ref{app: Implementation Details} and \ref{app:additional_results}. We also provide videos of the generated trajectories in the supplementary material.

\subsection{Comparison across Methods}
\begin{table*}[t]
    \centering
    \scriptsize
    \renewcommand{\arraystretch}{0.8}
    \setlength{\tabcolsep}{2pt}
    \resizebox{0.8\textwidth}{!}{%
    \begin{tabular}{lc|cccc|cccc|cccc|>{\columncolor{sidgray}}c>{\columncolor{sidgray}}c>{\columncolor{sidgray}}c>{\columncolor{sidgray}}c}
    \toprule
    Map & Robots
    & \multicolumn{4}{c|}{ORCA}
    & \multicolumn{4}{c|}{IA-MPC}
    & \multicolumn{4}{c|}{MIMIC-D}
    & \multicolumn{4}{c}{SID} \\
    \cmidrule(lr){3-6} \cmidrule(lr){7-10} \cmidrule(lr){11-14} \cmidrule(lr){15-18}
    & & S & C & A & T & S & C & A & T & S & C & A & T & S & C & A & T \\
    \midrule
    \multirow{3}{*}{\rotatebox[origin=c]{90}{\textbf{Basic}}}
    & 3  & 88 & 0 & 0.24 & 21.92 & 36 & 8 & 0.15 & 63.00 & 8 & 0 & 0.47 & 63.00 & \textbf{100} & 0.05 & 1.06 & 41.11 \\
    & 6  & 64 & 0 & 0.28 & 22.89 & 4  & 8 & 0.08 & 63.00 & 0 & 0 & N/A & N/A & \textbf{100} & 0.04 & 1.23 & 41.94 \\
    & 9  & 52 & 0 & 0.32 & 23.33 & 0  & 8 & N/A & N/A   & 0 & 0 & N/A & N/A & \textbf{100} & 0.07 & 1.03 & 42.35 \\
    \midrule
    \multirow{3}{*}{\rotatebox[origin=c]{90}{\textbf{Dense}}}
    & 3  & 96 & 0 & 0.26 & 23.35 & 20 & 8 & 0.16 & 63.00 & 12 & 0 & 0.65 & 63.00 & \textbf{100} & 0 & 1.17 & 41.45 \\
    & 6  & 64 & 0 & 0.30 & 24.27 & 0  & 8 & N/A & N/A   & 0 & 0 & N/A & N/A & \textbf{100} & 0.06 & 1.16 & 42.49 \\
    & 9  & 28 & 0 & 0.33 & 25.25 & 0  & 8 & N/A & N/A   & 0 & 0 & N/A & N/A & \textbf{100} & 0.06 & 1.12 & 42.48 \\
    \midrule
    \multirow{3}{*}{\rotatebox[origin=c]{90}{\textbf{Room}}}
    & 3  & 68 & 0 & 0.20 & 22.02 & 52 & 8 & 0.13 & 63.00 & 40 & 0 & 0.59 & 63.00 & \textbf{100} & 0.08 & 0.81 & 39.93 \\
    & 6  & 36 & 0 & 0.24 & 23.28 & 4  & 8 & 0.06 & 63.00 & 24 & 0 & 0.38 & 63.00 & \textbf{100} & 0.88 & 1.09 & 40.05 \\
    & 9  & 20 & 0 & 0.27 & 22.62 & 0  & 8 & N/A & N/A   & 12 & 0 & 0.26 & 63.00 & \textbf{84} & 3.08 & 0.97 & 39.92 \\
    \midrule
    \multirow{3}{*}{\rotatebox[origin=c]{90}{\textbf{Shelf}}}
    & 3  & 64 & 0 & 0.16 & 20.94 & 52 & 8 & 0.12 & 29.82 & 52 & 0 & 0.56 & 63.00 & \textbf{100} & 0.08 & 0.68 & 42.83 \\
    & 6  & 28 & 0 & 0.22 & 21.71 & 4  & 8 & 0.07 & 41.67 & 36 & 0 & 0.37 & 63.00 & \textbf{92} & 0.92 & 1.15 & 41.89 \\
    & 9  & 12 & 0 & 0.26 & 21.78 & 0  & 8 & N/A & N/A   & 8 & 0 & 0.24 & 63.00 & \textbf{96} & 1.12 & 1.02 & 41.29 \\
    \bottomrule
    \end{tabular}%
    }
    \caption{Results for the original benchmark robot counts. S denotes success rate (\%), C denotes communication frequency, A denotes acceleration-based smoothness, and T denotes travel time measured by average arrival time. Smoothness and travel time are calculated within feasible instances. Values in column A are reported in units of $10^{-2}$.}
    \label{tab:benchmark_original_robot_counts}
\end{table*}

Table~\ref{tab:benchmark_original_robot_counts} evaluates each method at the robot counts used in the original benchmarks.
\emph{ORCA} degrades significantly as robot count and map complexity increase. Specifically, its success drops from 88\% on Basic maps with three robots to 28\% on Dense maps, 20\% on Room maps, and 12\% on Shelf maps with nine robots. This drop reflects a limitation of reactive planning, which uses only the current local observation and does not anticipate neighboring robots.
\emph{IA-MPC} performs worse. Although it uses an RNN to predict neighboring robots' future trajectories as references for decentralized planning, large prediction errors can mislead the planner, yielding zero success with nine robots across all four map types.
\emph{MIMIC-D} uses diffusion models as decentralized planners, but generated trajectories are not guaranteed to satisfy planning constraints in complex environments. This limitation is consistent with observations in diffusion-based centralized planning~\citep{shaoul2024multi,liang2025simultaneous}. Because \emph{MIMIC-D} also plans from the current local observation, it cannot anticipate neighboring robots. It fails on Basic and Dense maps with nine robots and reaches only 12\% and 8\% success on Room and Shelf maps.

In contrast, SID achieves perfect success on Basic and Dense maps with up to nine robots, and reaches 84\% and 96\% success on Room and Shelf maps with nine robots. These results improve over the strongest baseline by \textbf{4.2$\times$} and \textbf{8.0$\times$} on the two structured maps.
\begin{wrapfigure}[19]{r}{0.7\textwidth}
    \centering
    \begin{subfigure}[t]{0.48\linewidth}
        \centering
        \includegraphics[width=\linewidth, trim={22 22 22 22}, clip]{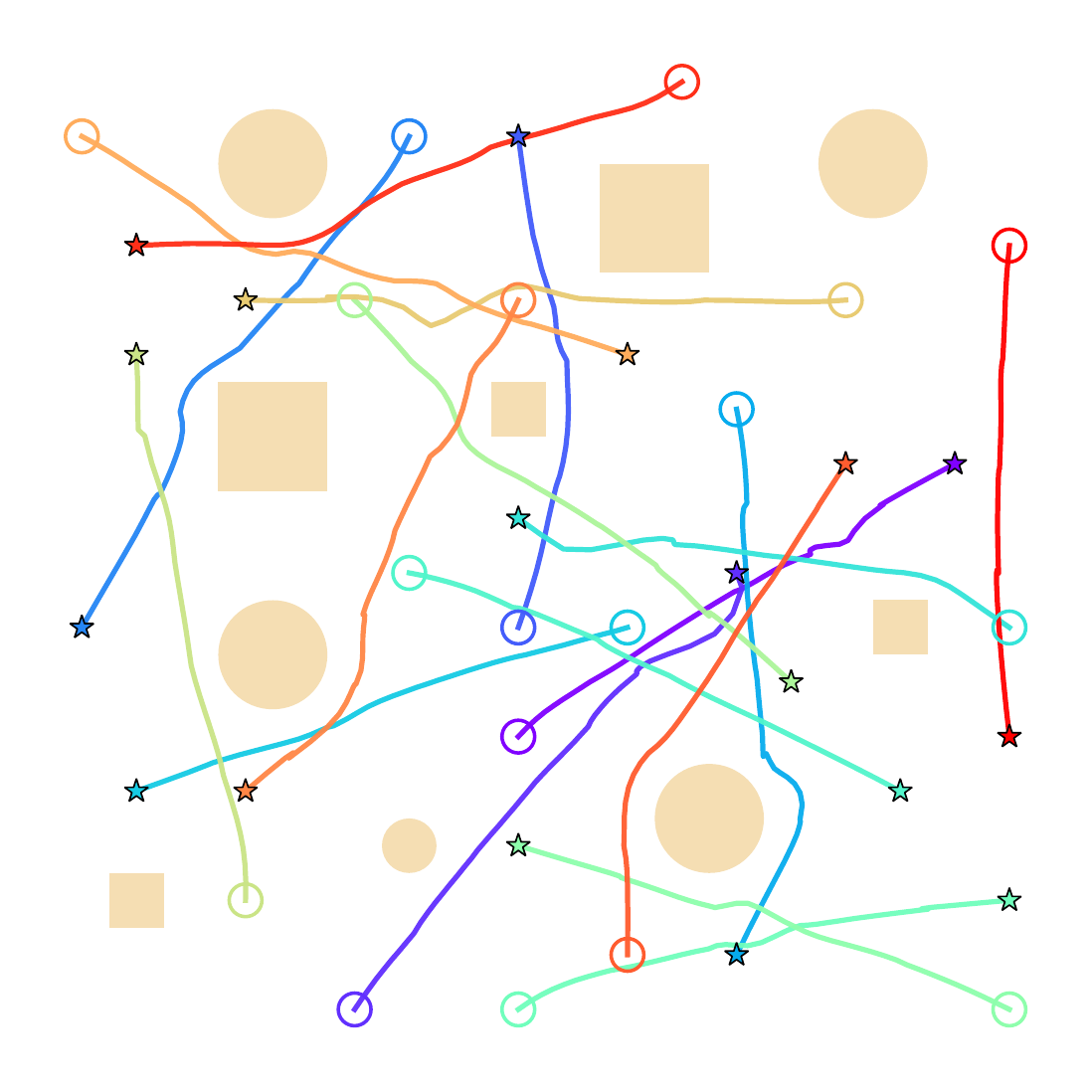}
        \caption{Basic map.}
        \label{fig:sid_18_basic}
    \end{subfigure}
    \hfill
    \begin{subfigure}[t]{0.48\linewidth}
        \centering
        \includegraphics[width=\linewidth, trim={22 22 22 22}, clip]{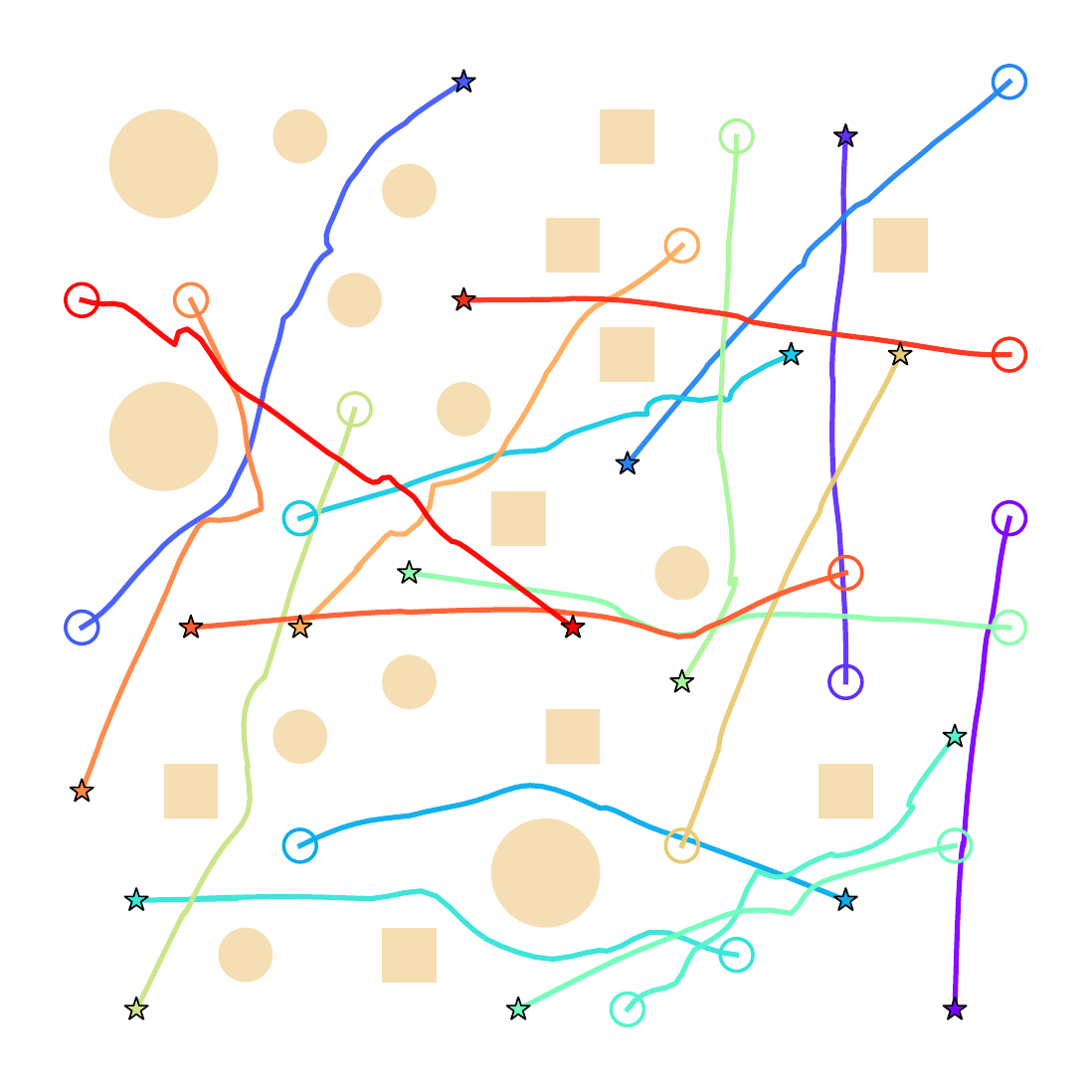}
        \caption{Dense map.}
        \label{fig:sid_18_dense}
    \end{subfigure}
    \caption{Trajectories generated by SID with 18 robots.}
    \label{fig:sid_18_robot_examples}
\end{wrapfigure}
SID also requires much less communication than IA-MPC. IA-MPC communicates before every planning round to collect neighboring trajectory histories, whereas SID simulates neighbors from the current local observation and communicates only when coordination is needed. Thus, SID's communication frequency stays close to that of communication-free methods.
SID has higher acceleration because acceleration is measured only on feasible trajectories, and harder successful cases often require more complex motions; the absolute values remain small. Finally, unlike the fixed-horizon arrival schedule used by IA-MPC and MIMIC-D, SID uses $\mathcal{C}_{\text{hold}}^\psi(i)$ to reduce per-robot travel time.
\begin{table*}[t]
    \centering
    \scriptsize
    \renewcommand{\arraystretch}{0.8}
    \setlength{\tabcolsep}{2pt}
    \resizebox{0.8\textwidth}{!}{%
    \begin{tabular}{lc|cccc|cccc|cccc|>{\columncolor{sidgray}}c>{\columncolor{sidgray}}c>{\columncolor{sidgray}}c>{\columncolor{sidgray}}c}
    \toprule
    Map & Robots
    & \multicolumn{4}{c|}{ORCA}
    & \multicolumn{4}{c|}{IA-MPC}
    & \multicolumn{4}{c|}{MIMIC-D}
    & \multicolumn{4}{c}{SID} \\
    \cmidrule(lr){3-6} \cmidrule(lr){7-10} \cmidrule(lr){11-14} \cmidrule(lr){15-18}
    & & S & C & A & T & S & C & A & T & S & C & A & T & S & C & A & T \\
    \midrule
    \multirow{3}{*}{\rotatebox[origin=c]{90}{\textbf{Basic}}}
    & 12 & 28 & 0 & 0.35 & 24.57 & 0  & 8 & N/A & N/A   & 0 & 0 & N/A & N/A & \textbf{100} & 0.08 & 1.12 & 42.48 \\
    & 15 & 4  & 0 & 0.37 & 25.93 & 0  & 8 & N/A & N/A   & 0 & 0 & N/A & N/A & \textbf{92}  & 0.20 & 0.92 & 42.21 \\
    & 18 & 0  & 0 & N/A & N/A   & 0  & 8 & N/A & N/A   & 0 & 0 & N/A & N/A & \textbf{80}  & 0.33 & 0.96 & 41.90 \\
    \midrule
    \multirow{3}{*}{\rotatebox[origin=c]{90}{\textbf{Dense}}}
    & 12 & 8  & 0 & 0.35 & 25.83 & 0  & 8 & N/A & N/A   & 0 & 0 & N/A & N/A & \textbf{92}  & 0.21 & 1.08 & 42.51 \\
    & 15 & 8  & 0 & 0.36 & 26.43 & 0  & 8 & N/A & N/A   & 0 & 0 & N/A & N/A & \textbf{96}  & 0.15 & 0.98 & 42.44 \\
    & 18 & 4  & 0 & 0.38 & 27.94 & 0  & 8 & N/A & N/A   & 0 & 0 & N/A & N/A & \textbf{88}  & 0.41 & 1.05 & 41.57 \\
    \bottomrule
    \end{tabular}%
    }
    \caption{Scalability results on the extended basic and dense maps.}
    \label{tab:benchmark_extended_robot_counts}
\end{table*}

Table~\ref{tab:benchmark_extended_robot_counts} shows that SID scales beyond the baselines as team size increases.
IA-MPC and MIMIC-D solve no extended cases. ORCA succeeds in only a few settings. As the team grows from 12 to 18 robots, ORCA drops from 28\% to 0\% on Basic maps and from 8\% to 4\% on Dense maps. SID maintains high success across 12, 15, and 18 robots: 100\%, 92\%, and 80\% on Basic maps, and 92\%, 96\%, and 88\% on Dense maps. At 18 robots, SID is the only successful method on Basic maps and exceeds ORCA by 84 percentage points on Dense maps. SID communicates sparsely, with frequency at most 0.41 compared with the fixed value of eight for IA-MPC. Figure~\ref{fig:sid_18_robot_examples} shows representative 18-robot trajectories generated by SID.

\subsection{Sensitivity Analysis}

\begin{figure}[t]
    \centering
    \includegraphics[width=0.8\textwidth]{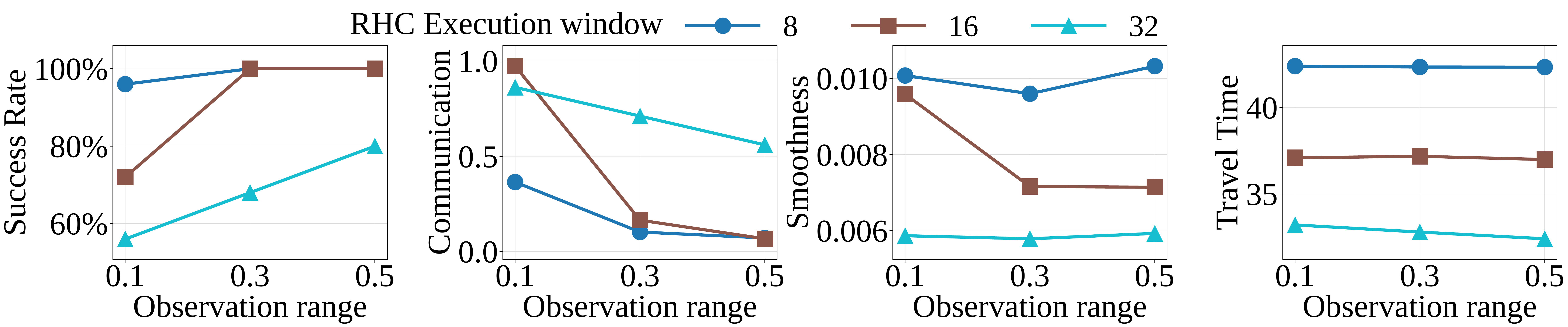}
    \caption{Sensitivity of SID on Basic maps with 9 robots. We sweep the local observation range and the execution window $k$ in receding-horizon control.}
    \label{fig:sensitivity_basic}
\end{figure}
Figure~\ref{fig:sensitivity_basic} studies the sensitivity of SID to two deployment parameters: the local observation range and the receding-horizon execution window $k$. The observation range primarily affects the quality of local interaction modeling. With a larger neighborhood, each robot can simulate more relevant neighbors, which improves success and reduces communication. At $k=16$, success increases from 72\% at observation range 0.1 to 100\% at ranges 0.3 and 0.5, while communication decreases from 0.97 to 0.16 and 0.07. The execution window $k$ controls the feedback frequency. Smaller windows replan more often and are therefore more robust to interaction uncertainty, while larger windows execute longer trajectory segments before replanning. In the tested settings, increasing $k$ from 8 to 32 reduces travel time from approximately 42.4 steps to 32.4--33.2 steps and reduces acceleration from approximately 0.010 to 0.006, but also lowers success from 96--100\% to 56--80\%. These trends show that the gains from longer execution windows are not free: they reduce feedback and can make the planner less robust. For the Basic-map benchmark with nine robots, $k=16$ with observation range 0.3 or 0.5 achieves the most balanced empirical behavior among the tested configurations, combining 100\% success, low communication, and shorter travel time than the more conservative $k=8$ setting.

\subsection{Ablation Studies}
\label{sec:ablation_studies}

To examine the roles of neighbor trajectory simulation and communication, we conduct ablation studies on the original benchmark and settings. Table~\ref{tab:ablation_studies} reports success rates for 3, 6, and 9 robots over 25 instances per setting.

\begin{table}[t]
    \centering
    \small
    \renewcommand{\arraystretch}{0.9}
    \setlength{\tabcolsep}{8pt}
    \begin{tabular}{l|ccc|ccc|ccc|ccc}
    \toprule
     Map & \multicolumn{3}{c|}{Basic} & \multicolumn{3}{c|}{Dense} & \multicolumn{3}{c|}{Room} & \multicolumn{3}{c}{Shelf} \\
    \cmidrule(lr){2-4} \cmidrule(lr){5-7} \cmidrule(lr){8-10} \cmidrule(lr){11-13}
    Robots & 3 & 6 & 9 & 3 & 6 & 9 & 3 & 6 & 9 & 3 & 6 & 9 \\
    \midrule
    NoS+NoC & 84 & 56 & 20 & 56 & 20 & 8 & 72 & 36 & 8 & 88 & 40 & 4 \\
    NoS+C & \textbf{100} & 76 & 28 & 80 & 44 & 16 & 88 & 60 & 20 & 88 & 56 & 4 \\
    S+NoC & 92 & 92 & 72 & \textbf{100} & 92 & 76 & 92 & 76 & 48 & 96 & 88 & 60 \\
    \rowcolor{sidgray}
    S+C (SID) & \textbf{100} & \textbf{100} & \textbf{100} & \textbf{100} & \textbf{100} & \textbf{100} & \textbf{100} & \textbf{100} & \textbf{84} & \textbf{100} & \textbf{92} & \textbf{96} \\
    \bottomrule
    \end{tabular}
    \caption{Ablation success rates (\%) on the original benchmark. \emph{S+C} is full SID; \emph{NoS+C} uses constant-velocity references with communication; \emph{S+NoC} retains CADM simulation but removes fallback communication; and \emph{NoS+NoC} uses CADM purely reactively from constant-velocity references.}
    \label{tab:ablation_studies}
\end{table}

With nine robots, SID achieves 100\% success on Basic and Dense maps and 84\% and 96\% on Room and Shelf maps, respectively. In contrast, reactive CADM (\emph{NoS+NoC}) achieves only 20\%, 8\%, 8\%, and 4\% across the same maps.  Even with communication, the constant-velocity variant (\emph{NoS+C}) achieves only 28\%, 16\%, 20\%, and 4\%, respectively. This means constant-velocity references may miss future conflicts, showing that communication alone cannot compensate for inaccurate neighbor references. Moreover, removing communication (\emph{S+NoC}) reduces success rates from SID's 100\%, 100\%, 84\%, and 96\% to 72\%, 76\%, 48\%, and 60\% on Basic, Dense, Room, and Shelf maps with nine robots, respectively, showing that simulation alone cannot resolve bottleneck deadlocks. These results indicate that simulation-informed planning contributes substantially to SID's performance gains beyond those provided by the CADM backbone or communication alone, while communication remains important for resolving coordination bottlenecks.

\subsection{Challenging Scenarios}

We further evaluate SID in large-scale environments that combine multiple obstacle patterns within a single map. These scenarios contain 108 robots and 160 obstacles, creating dense interactions and narrow passages. The robots have start--goal pairs with varying travel distances, requiring robots with short and long travel paths to coordinate under partial observations. To the best of our knowledge, no existing diffusion-based decentralized MRMP method has been demonstrated on environments of this scale and complexity.
Figure~\ref{fig:generated_trajectories_large} in Appendix~\ref{app:additional_results} shows all 108 trajectories alongside a view highlighting eight robots with longer travel paths.

\section{Conclusion}
\label{sec:conclusion}
This paper introduced Simulation-Informed Diffusion (SID), a decentralized MRMP framework that uses a constraint-aware diffusion model both to simulate neighboring robots and to plan each robot's own trajectory under safety constraints. By replacing snapshot-based reactive planning with simulated future references, SID accounts for spatial and temporal interactions over the planning horizon. The same simulation signal also supports a minimal communication mechanism that activates only at coordination bottlenecks where local planning becomes infeasible. Across diverse settings and challenging 108-robot scenarios, SID improves success rates over reactive, predictive, and diffusion-policy baselines while keeping communication sparse.\\
\textbf{Limitations and future work.} While SID provides state-of-the-art results in 2D decentralized MRMP tasks, several directions remain open. A natural next step is to extend the framework to 3D manipulation. Moreover, extending the proposed framework to settings where constraints are specified implicitly remains a promising direction for future work. Neighbor simulation also introduces additional computational overhead. Future work can consider faster constrained sampling.

\section*{Acknowledgment}
This research at the University of Virginia was partially supported by NSF grants 2334936, 2334448, NSF CAREER Award 2401285, and DARPA Contract No.~\#HR0011252E005.
The research at the University of California, Irvine was supported by NSF grants 2544613, 2434916, 2321786, 2112533, as well as gifts from Amazon Robotics and the Donald Bren Foundation. The authors acknowledge Research Computing at the University of Virginia for providing computational resources that have contributed to the results reported within this paper. The views and conclusions contained in this document are those of the authors and should not be interpreted as representing the official policies, either expressed or implied, of the sponsoring organizations, agencies, or the U.S. government. The authors declare no competing interests.

\newpage
\bibliographystyle{plainnat}
\bibliography{ref}  

@INPROCEEDINGS{6385823,
  author={Augugliaro, Federico and Schoellig, Angela P. and D'Andrea, Raffaello},
  booktitle={2012 IEEE/RSJ International Conference on Intelligent Robots and Systems}, 
  title={Generation of collision-free trajectories for a quadrocopter fleet: A sequential convex programming approach}, 
  year={2012},
  volume={},
  number={},
  pages={1917-1922}}

@inproceedings{jaitly2025milp,
  title={A MILP-based solution to multi-agent motion planning and collision avoidance in constrained environments},
  author={Jaitly, Akshay and Cline, Jack and Farzan, Siavash},
  booktitle={2025 IEEE 21st International Conference on Automation Science and Engineering (CASE)},
  pages={2200--2207},
  year={2025},
  organization={IEEE}
}

@article{marcucci2024shortest,
  title={Shortest paths in graphs of convex sets},
  author={Marcucci, Tobia and Umenberger, Jack and Parrilo, Pablo and Tedrake, Russ},
  journal={SIAM Journal on Optimization},
  volume={34},
  number={1},
  pages={507--532},
  year={2024},
  publisher={SIAM}
}

@article{sheng2006distributed,
  title={Distributed multi-robot coordination in area exploration},
  author={Sheng, Weihua and Yang, Qingyan and Tan, Jindong and Xi, Ning},
  journal={Robotics and autonomous systems},
  volume={54},
  number={12},
  pages={945--955},
  year={2006},
  publisher={Elsevier}
}

@INPROCEEDINGS{6942705,
  author={Wiktor, Adam and Scobee, Dexter and Messenger, Sean and Clark, Christopher},
  booktitle={2014 IEEE/RSJ International Conference on Intelligent Robots and Systems}, 
  title={Decentralized and complete multi-robot motion planning in confined spaces}, 
  year={2014},
  volume={},
  number={},
  pages={1168-1175},
  doi={10.1109/IROS.2014.6942705}}

@inproceedings{van2011reciprocal,
  title={Reciprocal n-body collision avoidance},
  author={Van Den Berg, Jur and Guy, Stephen J and Lin, Ming and Manocha, Dinesh},
  booktitle={Robotics research: the 14th international symposium ISRR},
  pages={3--19},
  year={2011},
  organization={Springer}
}

@ARTICLE{9384148,
  author={Guo, Ke and Wang, Dawei and Fan, Tingxiang and Pan, Jia},
  journal={IEEE Robotics and Automation Letters}, 
  title={VR-ORCA: Variable Responsibility Optimal Reciprocal Collision Avoidance}, 
  year={2021},
  volume={6},
  number={3},
  pages={4520-4527},
  doi={10.1109/LRA.2021.3067851}}

@inproceedings{alonso2013optimal,
  title={Optimal reciprocal collision avoidance for multiple non-holonomic robots},
  author={Alonso-Mora, Javier and Breitenmoser, Andreas and Rufli, Martin and Beardsley, Paul and Siegwart, Roland},
  booktitle={Distributed autonomous robotic systems: The 10th international symposium},
  pages={203--216},
  year={2013},
  organization={Springer}
}

@ARTICLE{8661608,
  author={Sartoretti, Guillaume and Kerr, Justin and Shi, Yunfei and Wagner, Glenn and Kumar, T. K. Satish and Koenig, Sven and Choset, Howie},
  journal={IEEE Robotics and Automation Letters}, 
  title={PRIMAL: Pathfinding via Reinforcement and Imitation Multi-Agent Learning}, 
  year={2019},
  volume={4},
  number={3},
  pages={2378-2385},
  doi={10.1109/LRA.2019.2903261}}

@ARTICLE{9366340,
  author={Damani, Mehul and Luo, Zhiyao and Wenzel, Emerson and Sartoretti, Guillaume},
  journal={IEEE Robotics and Automation Letters}, 
  title={PRIMAL$_2$: Pathfinding Via Reinforcement and Imitation Multi-Agent Learning - Lifelong}, 
  year={2021},
  volume={6},
  number={2},
  pages={2666-2673},
  doi={10.1109/LRA.2021.3062803}}

@INPROCEEDINGS{10342305,
  author={Wang, Yutong and Xiang, Bairan and Huang, Shinan and Sartoretti, Guillaume},
  booktitle={2023 IEEE/RSJ International Conference on Intelligent Robots and Systems (IROS)}, 
  title={SCRIMP: Scalable Communication for Reinforcement- and Imitation-Learning-Based Multi-Agent Pathfinding}, 
  year={2023},
  volume={},
  number={},
  pages={9301-9308},
  doi={10.1109/IROS55552.2023.10342305}}

@ARTICLE{9360433,
  author={Zhu, Hai and Claramunt, Francisco Martinez and Brito, Bruno and Alonso-Mora, Javier},
  journal={IEEE Robotics and Automation Letters}, 
  title={Learning Interaction-Aware Trajectory Predictions for Decentralized Multi-Robot Motion Planning in Dynamic Environments}, 
  year={2021},
  volume={6},
  number={2},
  pages={2256-2263},
  doi={10.1109/LRA.2021.3061073}}

@ARTICLE{10172259,
  author={Lindemann, Lars and Cleaveland, Matthew and Shim, Gihyun and Pappas, George J.},
  journal={IEEE Robotics and Automation Letters}, 
  title={Safe Planning in Dynamic Environments Using Conformal Prediction}, 
  year={2023},
  volume={8},
  number={8},
  pages={5116-5123},
  doi={10.1109/LRA.2023.3292071}}

@ARTICLE{11097366,
  author={Carvalho, João and Le, An Thai and Kicki, Piotr and Koert, Dorothea and Peters, Jan},
  journal={IEEE Transactions on Robotics}, 
  title={Motion Planning Diffusion: Learning and Adapting Robot Motion Planning With Diffusion Models}, 
  year={2025},
  volume={41},
  number={},
  pages={4881-4901},
  doi={10.1109/TRO.2025.3593109}}

@article{luo2024potential,
  title={Potential based diffusion motion planning},
  author={Luo, Yunhao and Sun, Chen and Tenenbaum, Joshua B and Du, Yilun},
  journal={arXiv preprint arXiv:2407.06169},
  year={2024}
}

@article{luan2025projected,
  title={Projected Coupled Diffusion for Test-Time Constrained Joint Generation},
  author={Luan, Hao and Goh, Yi Xian and Ng, See-Kiong and Ling, Chun Kai},
  journal={arXiv preprint arXiv:2508.10531},
  year={2025}
}

@article{shaoul2024multi,
  title={Multi-robot motion planning with diffusion models},
  author={Shaoul, Yorai and Mishani, Itamar and Vats, Shivam and Li, Jiaoyang and Likhachev, Maxim},
  journal={arXiv preprint arXiv:2410.03072},
  year={2024}
}

@InProceedings{liang2025simultaneous,
  title = 	 {Simultaneous Multi-Robot Motion Planning with Projected Diffusion Models},
  author =       {Liang, Jinhao and Christopher, Jacob K and Koenig, Sven and Fioretto, Ferdinando},
  booktitle = 	 {Proceedings of the 42nd International Conference on Machine Learning},
  pages = 	 {37162--37180},
  year = 	 {2025},
  volume = 	 {267},
  series = 	 {Proceedings of Machine Learning Research},
  month = 	 {13--19 Jul},
  publisher =    {PMLR},
}

@article{liang2025discrete,
  title={Discrete-guided diffusion for scalable and safe multi-robot motion planning},
  author={Liang, Jinhao and Koenig, Sven and Fioretto, Ferdinando},
  journal={arXiv preprint arXiv:2508.20095},
  year={2025}
}

@article{zhu2024madiff,
  title={Madiff: Offline multi-agent learning with diffusion models},
  author={Zhu, Zhengbang and Liu, Minghuan and Mao, Liyuan and Kang, Bingyi and Xu, Minkai and Yu, Yong and Ermon, Stefano and Zhang, Weinan},
  journal={Advances in Neural Information Processing Systems},
  volume={37},
  pages={4177--4206},
  year={2024}
}

@article{dong2025mimic,
  title={MIMIC-D: Multi-modal Imitation for MultI-agent Coordination with Decentralized Diffusion Policies},
  author={Dong, Dayi and Bhatt, Maulik and Choi, Seoyeon and Mehr, Negar},
  journal={arXiv preprint arXiv:2509.14159},
  year={2025}
}

@article{lew2025aid,
  title={AID: Agent Intent from Diffusion for Multi-Agent Informative Path Planning},
  author={Lew, Jeric and Cao, Yuhong and Tan, Derek Ming Siang and Sartoretti, Guillaume},
  journal={arXiv preprint arXiv:2512.02535},
  year={2025}
}

@article{ho2020denoising,
  title={Denoising diffusion probabilistic models},
  author={Ho, Jonathan and Jain, Ajay and Abbeel, Pieter},
  journal={Advances in neural information processing systems},
  volume={33},
  pages={6840--6851},
  year={2020}
}

@article{song2019generative,
  title={Generative modeling by estimating gradients of the data distribution},
  author={Song, Yang and Ermon, Stefano},
  journal={Advances in neural information processing systems},
  volume={32},
  year={2019}
}

@article{song2020score,
  title={Score-based generative modeling through stochastic differential equations},
  author={Song, Yang and Sohl-Dickstein, Jascha and Kingma, Diederik P and Kumar, Abhishek and Ermon, Stefano and Poole, Ben},
  journal={arXiv preprint arXiv:2011.13456},
  year={2020}
}

@ARTICLE{9384257,
  author={Chen, Yuxiao and Rosolia, Ugo and Ames, Aaron D.},
  journal={IEEE Robotics and Automation Letters}, 
  title={Decentralized Task and Path Planning for Multi-Robot Systems}, 
  year={2021},
  volume={6},
  number={3},
  pages={4337-4344},
  doi={10.1109/LRA.2021.3068103}}

@article{antonyshyn2023multiple,
  title={Multiple mobile robot task and motion planning: A survey},
  author={Antonyshyn, Luke and Silveira, Jefferson and Givigi, Sidney and Marshall, Joshua},
  journal={ACM Computing Surveys},
  volume={55},
  number={10},
  pages={1--35},
  year={2023},
  publisher={ACM New York, NY}
}

@InProceedings{pmlr-v155-wang21d,
  title = 	 {Model-based Reinforcement Learning for Decentralized Multiagent Rendezvous},
  author =       {Wang, Rose and Kew, J. Chase and Lee, Dennis and Lee, Tsang-Wei and Zhang, Tingnan and Ichter, Brian and Tan, Jie and Faust, Aleksandra},
  booktitle = 	 {Proceedings of the 2020 Conference on Robot Learning},
  pages = 	 {711--725},
  year = 	 {2021},
  editor = 	 {Kober, Jens and Ramos, Fabio and Tomlin, Claire},
  volume = 	 {155},
  series = 	 {Proceedings of Machine Learning Research},
  month = 	 {16--18 Nov},
  publisher =    {PMLR},
  url = 	 {https://proceedings.mlr.press/v155/wang21d.html}
}

@article{neal2011distributed,
  title={Distributed optimization and statistical learning via the alternating direction method of multipliers},
  author={Neal, Parikh and Eric, Chu and Borja, Peleato and Jonathan, Eckstein},
  journal={Foundations and Trends{\textregistered} in Machine learning},
  volume={3},
  number={1},
  pages={1--122},
  year={2011},
  publisher={Emerald Publishing Limited}
}

@book{boyd2004convex,
  title={Convex optimization},
  author={Boyd, Stephen and Vandenberghe, Lieven},
  year={2004},
  publisher={Cambridge university press}
}

@article{liang2026improved,
  title={Improved Constrained Generation by Bridging Pretrained Generative Models},
  author={Liang, Xiaoxuan and Naderiparizi, Saeid and Liu, Yunpeng and Zwartsenberg, Berend and Wood, Frank},
  journal={arXiv preprint arXiv:2603.06742},
  year={2026}
}

@inproceedings{silver2005cooperative,
  title={Cooperative pathfinding},
  author={Silver, David},
  booktitle={Proceedings of the aaai conference on artificial intelligence and interactive digital entertainment},
  volume={1},
  pages={117--122},
  year={2005}
}

@inproceedings{Fioretto:NeurIPS24,
  title = {Constrained Synthesis with Projected Diffusion Models},
  author = {Christopher, Jacob K and Baek, Stephen and Fioretto, Ferdinando},
  booktitle = {Advances in Neural Information Processing Systems},
  year = {2024},
  volume = {37},
  pages = {},
  publisher = {Curran Associates, Inc.},
}

@article{cen2025worldvla,
  title={Worldvla: Towards autoregressive action world model},
  author={Cen, Jun and Yu, Chaohui and Yuan, Hangjie and Jiang, Yuming and Huang, Siteng and Guo, Jiayan and Li, Xin and Song, Yibing and Luo, Hao and Wang, Fan and others},
  journal={arXiv preprint arXiv:2506.21539},
  year={2025}
}

@article{li2026causal,
  title={Causal world modeling for robot control},
  author={Li, Lin and Zhang, Qihang and Luo, Yiming and Yang, Shuai and Wang, Ruilin and Han, Fei and Yu, Mingrui and Gao, Zelin and Xue, Nan and Zhu, Xing and others},
  journal={arXiv preprint arXiv:2601.21998},
  year={2026}
}

@article{ye2026world,
  title={World action models are zero-shot policies},
  author={Ye, Seonghyeon and Ge, Yunhao and Zheng, Kaiyuan and Gao, Shenyuan and Yu, Sihyun and Kurian, George and Indupuru, Suneel and Tan, You Liang and Zhu, Chuning and Xiang, Jiannan and others},
  journal={arXiv preprint arXiv:2602.15922},
  year={2026}
}

@article{ma2026posterior,
  title={Posterior Flow Matching},
  author={Ma, Chenrui and Xiao, Xi and Zhao, Lin and Jiang, Xinru and Wang, Tianyang and Shen, Yanning and Fioretto, Ferdinando},
  journal={arXiv preprint arXiv:2610.13994},
  year={2026}
}

@inproceedings{ma2026learning,
  title={Learning straight flows: Variational flow matching for efficient generation},
  author={Ma, Chenrui and Xiao, Xi and Wang, Tianyang and Wang, Xiao and Shen, Yanning},
  booktitle={Proceedings of the IEEE/CVF Conference on Computer Vision and Pattern Recognition},
  pages={38154--38164},
  year={2026}
}

@article{ma2026drift,
  title={Drift Flow Matching},
  author={Ma, Chenrui and Xiao, Xi and Zhao, Lin and Wang, Tianyang and Fioretto, Ferdinando and Shen, Yanning},
  journal={arXiv preprint arXiv:2605.17244},
  year={2026}
}

@article{ma2026transition,
  title={Transition flow matching},
  author={Ma, Chenrui},
  journal={arXiv preprint arXiv:2603.15689},
  year={2026}
}

\newpage
\appendix

\section{Implementation Details}
\label{app: Implementation Details}

\subsection{Software and Hardware}
Software used for experiments is Ubuntu 22.04.5, Python 3.8.20, CUDA 12.1, and PyTorch 2.1.2. For each of our experiments, we used 1 NVIDIA RTX A6000 GPU.

\subsection{Training Details}
Our implementation builds upon the official code of~\citet{liang2025simultaneous} and~\citet{shaoul2024multi}, with modifications to accommodate our specific requirements. Table~\ref{table: hyperparams for Training} shows the hyperparameters used for training diffusion models in our experiments. 
\begin{table}[h]
\centering
\caption{Hyperparameters for Training in Experiments.}
\begin{tabular}{cc}
\hline
HyperParameters         & Value \\ \hline
Diffusion Sampling Step &   25    \\
Learning Rate           &   1e-4   \\
Batch Size              &   64   \\
Optimizer               &   Adam    \\ \hline
\end{tabular}
\label{table: hyperparams for Training}
\end{table}

\subsection{Planning Details.}
All experiments are conducted on $2 \times 2$ maps. The robot radius was set to 0.03 units across four different maps. Other parameters related to planning are summarized in Table~\ref{tab:app-impl-params}. 

\begin{table}[h]
\centering
\small
\caption{Simulation and planning parameters used in our experiments.}
\label{tab:app-impl-params}
\begin{tabular}{ll}
\toprule
Parameter & Value \\
\midrule
Map size & $2 \times 2$ units \\
Robot radius & $0.03$ units \\
Maximum speed & $v_{\max}=0.05$ \\
Maximum Planning horizon & $H=64$ \\
Receding-horizon execution window & $k=8$ \\
Receding-horizon execution window in sensitivity analysis & $k=8, 16, 32$ \\
Maximum observation range & $0.5 \times \mathrm{map \ size}$ \\
Maximum observation range in sensitivity analysis & $0.5, 0.3, 0.1 \times \mathrm{map \ size}$ \\
\bottomrule
\end{tabular}
\end{table}

\subsection{Benchmark Details.}
The benchmark consists of four map families: basic, dense, room, and shelf. 
Each type of map contains 25 test instances, where each instance specifies both the obstacle layout and the start--goal pairs. 
The map configurations are summarized in Table~\ref{tab:app-map-families}. 
For the standard benchmark, we evaluate 3, 6, and 9 robots on all four maps, resulting in $4 \times 3 \times 25 = 300$ test instances per method. 
For the extended scalability benchmark, we additionally evaluate 12, 15, and 18 robots on the basic and dense maps, resulting in another $2 \times 3 \times 25 = 150$ test instances per method.

\begin{table}[h]
\centering
\small
\caption{Map families used in the benchmark. Each map family contains 25 test instances.}
\label{tab:app-map-families}
\begin{tabular}{lll}
\toprule
Map family & Obstacle configuration & Obstacle size \\
\midrule
Basic & 10 randomly placed obstacles & $0.1$--$0.2$ units \\
Dense & 20 randomly placed obstacles & $0.1$--$0.2$ units \\
Room & 2 long rectangular obstacles & $0.2 \times 1.0$ units \\
Shelf & 2 long rectangular obstacles & $0.2 \times 1.0$ units \\
\bottomrule
\end{tabular}
\end{table}

\subsection{Evaluation Metrics.}
\paragraph{Success rate.}
An instance is counted as successful only if all robots reach their goals and the executed trajectories are collision-free with respect to both environmental obstacles and other robots. The reported success rate is
\begin{equation}
S =
100\cdot
\frac{\mathrm{number \ of \ successful \ instances}}{\mathrm{total \ number \ of \ instances}}
\end{equation}

\paragraph{Communication frequency.}
For SID, communication is activated only when the simulation-guided planner
cannot find feasible solutions. The instance-level communication frequency is
\begin{equation}
C_q = \frac{\mathrm{total \ number \ of \ communication \ events}}{\mathrm{number \ of \ robots}}
\end{equation}
and the reported communication frequency is
\begin{equation}
C =
\frac{\sum_1^Q C_q}{\mathrm{total \ number \ of \ instances}}.
\end{equation}
In the implementation, ORCA and MIMIC-D use no explicit communication, so their communication frequency is zero. IA-MPC needs to communicate at every receding-horizon planning round; with $H=64$ and $k=8$, this yields a fixed communication frequency of $H/k=8$.

\paragraph{Smoothness.}
Smoothness is measured as the integral of squared discrete acceleration. For robot $i$ in instance $q$, we compute
\begin{equation}
A_{i,q}
=
\Delta t
\sum_{t=0}^{H-2}
\left\|
\frac{
\mathbf{p}_{i,q}^{t+2}
-2\mathbf{p}_{i,q}^{t+1}
+\mathbf{p}_{i,q}^{t}
}{\Delta t^2}
\right\|_2^2 .
\end{equation}
The instance-level smoothness is the mean across robots,
\begin{equation}
A_q = \frac{1}{N}\sum_{i=1}^{N} A_{i,q}.
\end{equation}
We report the mean smoothness over successful instances. In the result tables, smoothness values are reported in units of $10^{-2}$.

\paragraph{Travel time.}
For instance $q$, we define travel time as the average arrival time across robots. We report the mean travel time over successful instances.


\section{Additional Experimental Results}
\label{app:additional_results}

\begin{table*}[b]
    \centering
    \scriptsize
    \renewcommand{\arraystretch}{0.8}
    \setlength{\tabcolsep}{3pt}
    \resizebox{0.8\textwidth}{!}{%
    \begin{tabular}{l|ccc|ccc|ccc|>{\columncolor{sidgray}}c>{\columncolor{sidgray}}c>{\columncolor{sidgray}}c}
    \toprule
    \textbf{Map}
    & \multicolumn{3}{c|}{\textbf{ORCA}}
    & \multicolumn{3}{c|}{\textbf{IA-MPC}}
    & \multicolumn{3}{c|}{\textbf{MIMIC-D}}
    & \multicolumn{3}{>{\columncolor{sidgray}}c}{\textbf{SID}} \\
    \cmidrule(lr){2-4}
    \cmidrule(lr){5-7}
    \cmidrule(lr){8-10}
    \cmidrule(lr){11-13}
    & \textbf{3} & \textbf{6} & \textbf{9}
    & \textbf{3} & \textbf{6} & \textbf{9}
    & \textbf{3} & \textbf{6} & \textbf{9}
    & \textbf{3} & \textbf{6} & \textbf{9} \\
    \midrule
    Basic
    & 0.11 & 0.35 & 0.68
    & 48.24 & 38.45 & N/A
    & 7.52 & N/A & N/A
    & 76.01 & 125.04 & 175.92 \\
    
    Dense
    & 0.34 & 0.87 & 1.75
    & 115.50 & N/A & N/A
    & 4.89 & N/A & N/A
    & 140.03 & 205.73 & 261.74 \\
    
    Room
    & 0.09 & 0.26 & 0.39
    & 54.71 & 29.27 & N/A
    & 19.99 & 36.80 & 52.33
    & 67.58 & 294.04 & 381.67 \\
    
    Shelf
    & 0.10 & 0.29 & 0.51
    & 58.96 & 33.01 & N/A
    & 19.37 & 28.24 & 51.41
    & 94.24 & 406.32 & 552.94 \\
    \bottomrule
    \end{tabular}%
    }
    \caption{Running time on the original benchmark robot counts. Values are reported in seconds and averaged over successful instances only. N/A indicates that the method has no successful instance under the corresponding setting.}
    \label{tab:running_time_original_robot_counts}
\end{table*}
This section provides additional results that complement the main experimental evaluation. Table~\ref{tab:running_time_original_robot_counts} reports the average running time of each method on successful instances. This should be interpreted together with the success rates in Table~\ref{tab:benchmark_original_robot_counts}, since running time is averaged only over instances that are successfully solved. Specifically, methods with lower success rates may exhibit lower reported running times, as they tend to succeed primarily on easier instances that can be solved quickly.
ORCA is the fastest method because it relies on lightweight reactive velocity selection, but its success rate decreases substantially in complex environments. IA-MPC and MIMIC-D often fail to solve any instances in harder settings. In particular, the running time of IA-MPC decreases as the number of robots increases on basic maps because the few instances it solves are comparatively simple. SID achieves the highest success rates while maintaining reasonable running times.

\begin{figure}[t]
    \centering
    \begin{subfigure}[t]{0.49\textwidth}
        \centering
        \includegraphics[width=\textwidth]{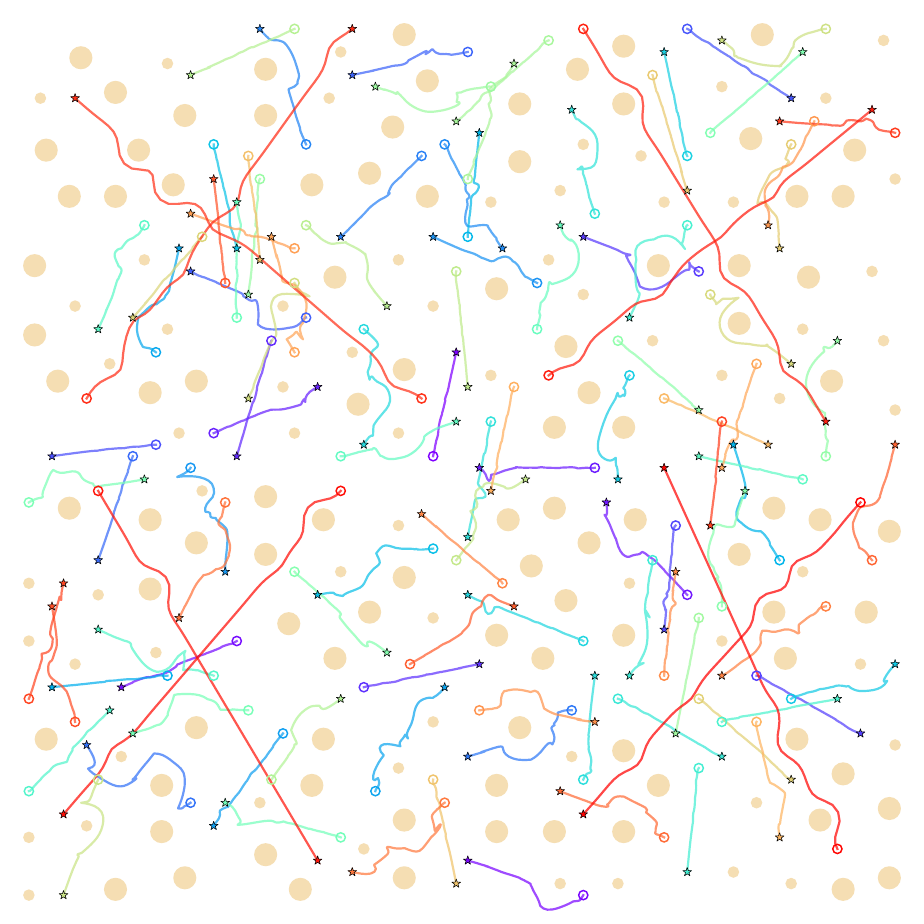}
        \caption{All 108 generated trajectories.}
        \label{fig:generated_trajectories_large_all}
    \end{subfigure}
    \hfill
    \begin{subfigure}[t]{0.49\textwidth}
        \centering
        \includegraphics[width=\textwidth]{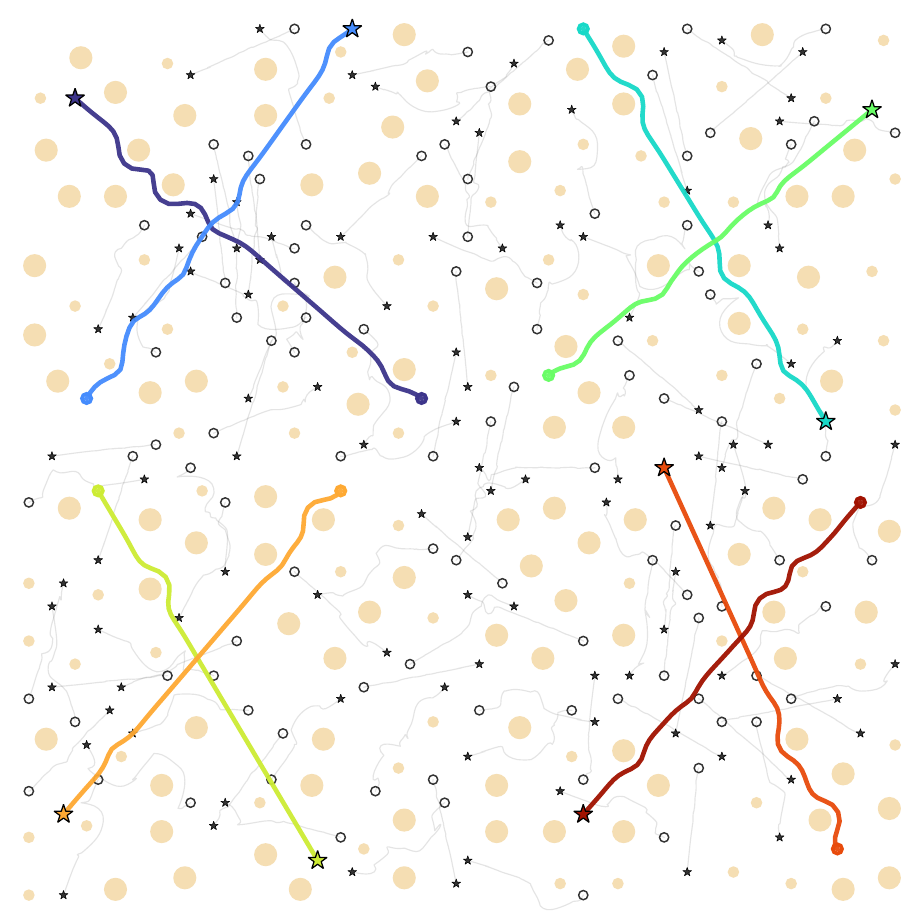}
        \caption{Trajectories of 8 robots with longer travel paths.}
        \label{fig:generated_trajectories_large_highlight}
    \end{subfigure}
    \caption{Representative trajectories generated by SID in a large-scale scenario with 108 robots and 160 obstacles.
    Left: all generated trajectories. Right: eight robots with longer travel paths are highlighted.}
    \label{fig:generated_trajectories_large}
\end{figure}

Figure~\ref{fig:generated_trajectories_large} further visualizes SID in a large-scale scenario with 108 robots. The generated trajectories show that SID can coordinate a great number of robots through cluttered regions and narrow passages, illustrating the scalability of the proposed simulation-guided planning and communication mechanism beyond the standard benchmark sizes. The start--goal pairs span different travel distances, so robots with short and long travel paths must coordinate within the same environment. Figure~\ref{fig:generated_trajectories_large_all} shows all 108 trajectories, while Figure~\ref{fig:generated_trajectories_large_highlight} highlights eight robots with longer travel paths. 

\end{document}